\documentclass[10pt,twocolumn,letterpaper]{article}

\usepackage{cvpr}              %

\usepackage[dvipsnames]{xcolor}

\usepackage{pifont}%
\newcommand{\cmark}{\ding{51}}%
\newcommand{\xmark}{\ding{55}}%

\newcommand{\vv}{\mathbf{v}}

\newcommand{\vS}{\mathbf{S}}

\newcommand{\vB}{\mathbf{B}}

\newcommand{\vt}{\mathbf{t}}

\definecolor{namebluecolor}{RGB}{76, 174, 227}
\definecolor{nameredcolor}{RGB}{250, 88, 51}%

\definecolor{cvprblue}{rgb}{0.21,0.49,0.74}
\usepackage[pagebackref,breaklinks,colorlinks,citecolor=cvprblue]{hyperref}
\usepackage[framemethod=TikZ]{mdframed}
\usepackage{makecell}
\usepackage{tikz}
\usepackage{xcolor}
\definecolor{lightblue}{rgb}{0.4,0.78039,1}
\usetikzlibrary{shapes, arrows.meta, positioning}

\def\paperID{15851} %
\def\confName{CVPR}
\def\confYear{2024}

\title{Zero-shot Referring Expression Comprehension via Structural Similarity Between Images and Captions}

\author{Zeyu Han\(^1\),~~ Fangrui Zhu\(^1\),~~ Qianru Lao\(^2\),~~ Huaizu Jiang\(^1\)\\
\(^1\)Northeastern University \quad \(^2\)Harvard University\\
{\tt\small \{han.zeyu,zhu.fang,h.jiang\}@northeastern.edu, estherbear17@gmail.com}
}
\begin{document}
\maketitle
\begin{abstract}
\textbf{Zero-shot referring expression comprehension} aims at localizing bounding boxes in an image corresponding to provided textual prompts, which requires: (i) a fine-grained disentanglement of complex visual scene and textual context, and (ii) a capacity to understand relationships among disentangled entities. Unfortunately, existing large vision-language alignment (VLA) models, \eg, CLIP, struggle with both aspects so cannot be directly used for this task. To mitigate this gap, we leverage large foundation models to disentangle both images and texts into triplets in the format of \((\texttt{subject}, \texttt{predicate}, \texttt{object})\). After that, grounding is accomplished by calculating the structural similarity matrix between visual and textual triplets with a VLA model, and subsequently propagate it to an instance-level similarity matrix. Furthermore, to equip VLA models with the ability of relationship understanding, we design a triplet-matching objective to fine-tune the VLA models on a collection of curated dataset containing abundant entity relationships. 
Experiments demonstrate that our visual grounding performance increase of up to 19.5\% over the SOTA zero-shot model on RefCOCO/+/g. On the more challenging Who's Waldo dataset, our zero-shot approach achieves comparable accuracy to the fully supervised model. Code is available at \url{https://github.com/Show-han/Zeroshot_REC}.
\end{abstract}
    
\section{Introduction}
\label{sec:intro}
Visual grounding is a fundamental task across computer vision and natural language processing, where the goal is to find the correspondences between image content and textual descriptions. It has broad applications in image captioning \cite{captionsurvey1, captionsurvey2}, visual question answering \cite{vqa1, vqa2}, vision-language navigation \cite{vision-navigation}, etc.
Collecting detailed grounding annotations to train specialist models, however, is cumbersome.
Therefore, \textbf{zero-shot visual grounding} \cite{reclip, diffvg, clip-phrase-grounding} is an attractive alternative.

\begin{figure}[t]
    \centering
    \includegraphics[width=0.9\linewidth]{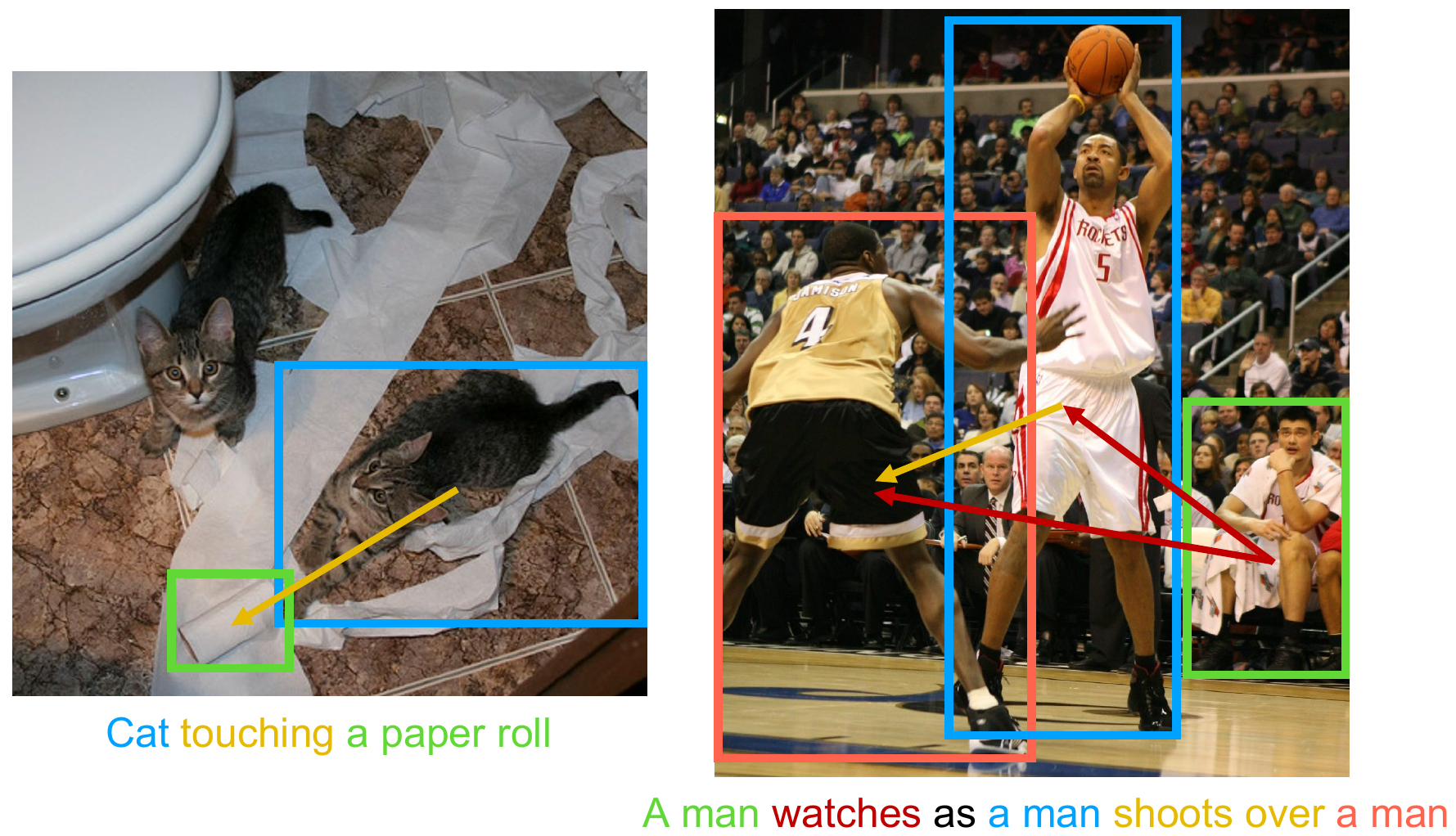}
    \caption{\textbf{Illustration of how we disambiguate visual entities based on their interactions with other entities.} The same entity or relationships in the image and caption are in the same color.
    }
    \vspace{-8pt}
    \label{fig:teaser}
\end{figure}

As a visual grounding task, the essence of referring expression comprehension (REC) is the alignment of text queries with corresponding image regions. 
To achieve this goal, it is critical for a grounding model to understand the relationships of entities\cite{visual-relationships}, both within and cross different modalities (visual \emph{vs.} textual), when identifying referred entities within an image. 
As shown in the upper part of Fig.~\ref{fig:teaser}, the \texttt{touching} relationship is the key to resolve the ambiguity of identifying the correct \texttt{cat}.
Similarly, the lower part of Fig.~\ref{fig:teaser} illustrates a more complex situation where multiple entities are engaged in various interactions. 
Both scenarios highlight the importance of \emph{relationship understanding within both the image and caption}, where entities are not merely isolated elements but interact dynamically with others in the scene. 
In a zero-shot learning context, the task of understanding these relationships can be more challenging, as the model lacks exposure to specific training instances that could aid in interpretation.

Recent advances in zero-shot REC \cite{reclip, clip-phrase-grounding, zero-shot-seg} have been largely driven by the integration of large-scale vision-language aligned (VLA) models such as CLIP \cite{clip} and FLAVA \cite{flava}, which serve as bridges connecting text and image domain. 
These approaches, however, fall short in relationship understanding.
On the one hand, in the textual domain, existing approaches adopt hand-crafted language parsers~\cite{reclip,zero-shot-seg} to decompose the input caption into a set of phrases, which are fragile and do not generalize well to long, complex captions in real-world applications, as shown in the bottom of Fig.~\ref{fig:teaser}.
On the other hand, the visual relationship understanding capability of VLA models is inherently not good enough. 
Recent studies have revealed that VLA models behave like ``bags-of-words'' \cite{bows}, and demonstrated that they fail to perform beyond chance level at simple tasks requiring compositional understanding \cite{bag-of-region, bows, dac, structured-vl, visual-programming}. 
Some effort have been dedicated to mitigate this issue by generating hard negative prompts through word replacement \cite{structure-CLIP, bows, structured-vl}, caption augmentation \cite{dac, structured-vl}, and feature augmentation \cite{comclip}. These rule-based methods, however, are limited in producing diverse samples and have potential bias of design pattern, consequently restricting their generalization capabilities.

In this paper, we focus on the REC task by explicitly modeling the entity relations within both images and captions using 
the \emph{structural similarity} between them to solve the zero-shot visual grounding problem. 
Specifically, we decompose the image and caption into two sets of triplets in the form of \((\texttt{subject}, \texttt{predicate}, \texttt{object})\), in which each triplet captures a pair of potential entities with their interrelation. 
By considering the similarity of \texttt{subject}, \texttt{object}, and \texttt{predicate} jointly, we can find better matchings of object proposals and their referrings.
Compared with existing work~\cite{reclip}, our approach is more principled and eliminates the ad-hoc post-processing spatial relation resolver.
More importantly, to improve the relationship understanding in the caption, we resort to ChatGPT and leverage its powerful \emph{in-context learning} capability \cite{gentopia, incontext} for the triplet decomposition to find all possible relation triplets given a sentence.
In contrast to the dependency parser in~\cite{reclip}, our parsing works better when dealing with long captions and do not restrict to spatial relationships (\eg., \texttt{to the left of}), which can fully capture the rich compositional semantics present in actions or interactions, such as \texttt{walking} or \texttt{talking to}.

To address the limitation of a VLA model's visual relationship understanding, we harness a curated collection of data sources rich in relational knowledge, which include human-object interaction datasets \cite{hico, swig} and image scene graph dataset \cite{vg}.
Similar to our grounding pipeline, we isolate visual entities and construct triplets in both visual and textual sides and then implement a triplet-level contrastive learning objective to fine-tune the VLA model.
Compared with the existing rule-based negative prompts construction, this design has two unique advantages. 
First, by decomposing a single image into multiple triplets, we can obtain more training instances and improve the diversity of the training data than simply using the entire image for fine-tuning.
Furthermore, 
the isolation of entities removes the distraction of other content in the image, providing more useful supervision for the model fine-tuning.
We fine-tune a VLA model in a parameter-efficient manner~\cite{peft_survey} using LoRA \cite{lora}, improving its visual relationship understanding while preserving its powerful generic feature representations learned from large-scale data.
Our resulting model is called \textbf{VR-VLA} (\textbf{V}isual \textbf{R}elationship VLA).

Experimental results show that on the standard RefCOCO/g/+ datasets \cite{refcoco1, refcoco2}, we can surpass the SOTA zero-shot baseline \cite{reclip} up to 19.5\%, and an average of 9.7\%. 
On the challenging Who's Waldo dataset \cite{whos-waldo}, whose captions are much longer and depict much richer interactions of humans, our zero-shot method significantly outperforms~\cite{reclip}, achieving comparable accuracy to supervised methods.
We also conduct ablation studies validating the effectiveness of our model design.

To summarize, our main contributions are three-folded. (1) A novel zero-shot visual grounding model, where we harness the powerful capabilities of foundation models (\eg, ChatGPT and CLIP) to explicitly model the structural similarity between entities. (2) A novel recipe to improve a VLA model's visual relationship understanding by efficiently incorporating the supervision from a collection of curated data sources. (3) We report SOTA zero-shot visual grounding results on the REC datasets and also show promising results on the Who's Waldo dataset, where our zero-shot approach achieves comparable accuracy to the fully supervised method.

\section{Related Work}
\label{sec:related_work}
\noindent\textbf{Visual grounding.}
Based on the focus of the grounding task, visual grounding diverges into two main categories. The first emphasizes the noun properties of the query text. Precise understanding of the noun's meaning enables locating the corresponding grounding box in the image. Representative datasets contain MS-COCO \cite{mscoco}, Object365 \cite{object365} (fixed-category), Flickr30K Entities \cite{flickr30k} (open-vocabulary), etc. The second category emphasizes comprehending the interrelations among entities to localize the correct visual entity (potentially from among many similar ones) corresponding to the query text. Representative datasets contain RefCOCO/+/g \cite{refcoco1, refcoco2}, Who's Waldo \cite{whos-waldo}, etc. Our research of interest belongs to the second category, where relations are strong grounding clues.

Based on the use of grounding data, visual grounding can be classified into supervised and zero-shot categories. The majority of research \cite{glipv2, mdetr, uninext} focuses on supervised visual grounding, where models are specifically designed and trained with grounding data for this purpose. Conversely, zero-shot visual grounding methods \cite{reclip, clip-phrase-grounding, diffvg} adapt pre-existing vision-language models for grounding tasks. Our work is situated within this zero-shot visual grounding paradigm.

\noindent\textbf{Visual relationship understanding.} 
Images and texts are constructed using fundamental elements~—~objects in images and noun spans in texts~—~along with their interactions and relationships. For a model to understand relationship, it must not only detect individual entities but also establish relational links between them. Research communities such as human-object interaction \cite{hico, swig, hoi_drg20,hoi_scg21,hoi_upt22,hoi_stip22,hoi_cdn21,hoi_hotr21,hoi_genvlkt22,hoi_fgahoi23, hoi_muren23,zhu2023diagnosing} and visual scene graph \cite{vg, psg,xu2017scene,li2017scene,zellers2018neural,tang2019learning,zhang2017ppr,zhong2021learning} emphasize the relational aspect for visual tasks. 

Visual relationship understanding is also highly relevant to the \emph{compositional reasoning} \cite{bows, structure-CLIP, clipscenegraph, visual-programming, structured-vl, comclip,fan2024muffin} ability for VLA models. Although we anticipate that VLA models, trained via contrastive learning on extensive image-text pairs, would inherently develop a capacity for compositional reasoning, the reality is somewhat different. Most SOTA VLA models behave like ``bags-of-words'' \cite{bows}. They are capable of matching textual entities with corresponding visual elements, but falling short in interpreting their relationships or attributes. To address this, many studies have implemented strategies such as introducing negative text prompts \cite{dac, bows, structured-vl} in training batches~—~including noun substitutions, attribute and verb modifications, caption augmentation~—~or integrating scene graphs \cite{clipscenegraph} during training or inference. In our work, we show that fine-tune the VLA models on the visual relationship datasets can alleviate this problem. Furthermore, the explicitly defined structural representation also help strengthen the compositional reasoning for visual grounding tasks.

\noindent\textbf{Language parsing.} In the language side, structure prediction~\cite{bertsrl,christou2021improving,belanger2016structured,paolini2021structured} is well studied and aims for solving several problems including entity recognition~\cite{entity-recognition1, entity-recognition2, entity-recognition3,gu2020data}, relation classification~\cite{relationcls1, relationcls2}, semantic role labeling~\cite{srl1, srl2}, event extraction~\cite{eventext1, eventext2}, coreference resolution~\cite{corefres1, corefres2}, etc. The acquired structural representation can be illustrated as either a language parsing tree~\cite{reclip} or a set of labels that indicate the respective roles of each word~\cite{bertsrl}.

\section{Proposed Approach}

\begin{figure}[t]
    \centering
    \includegraphics[width=1\linewidth]{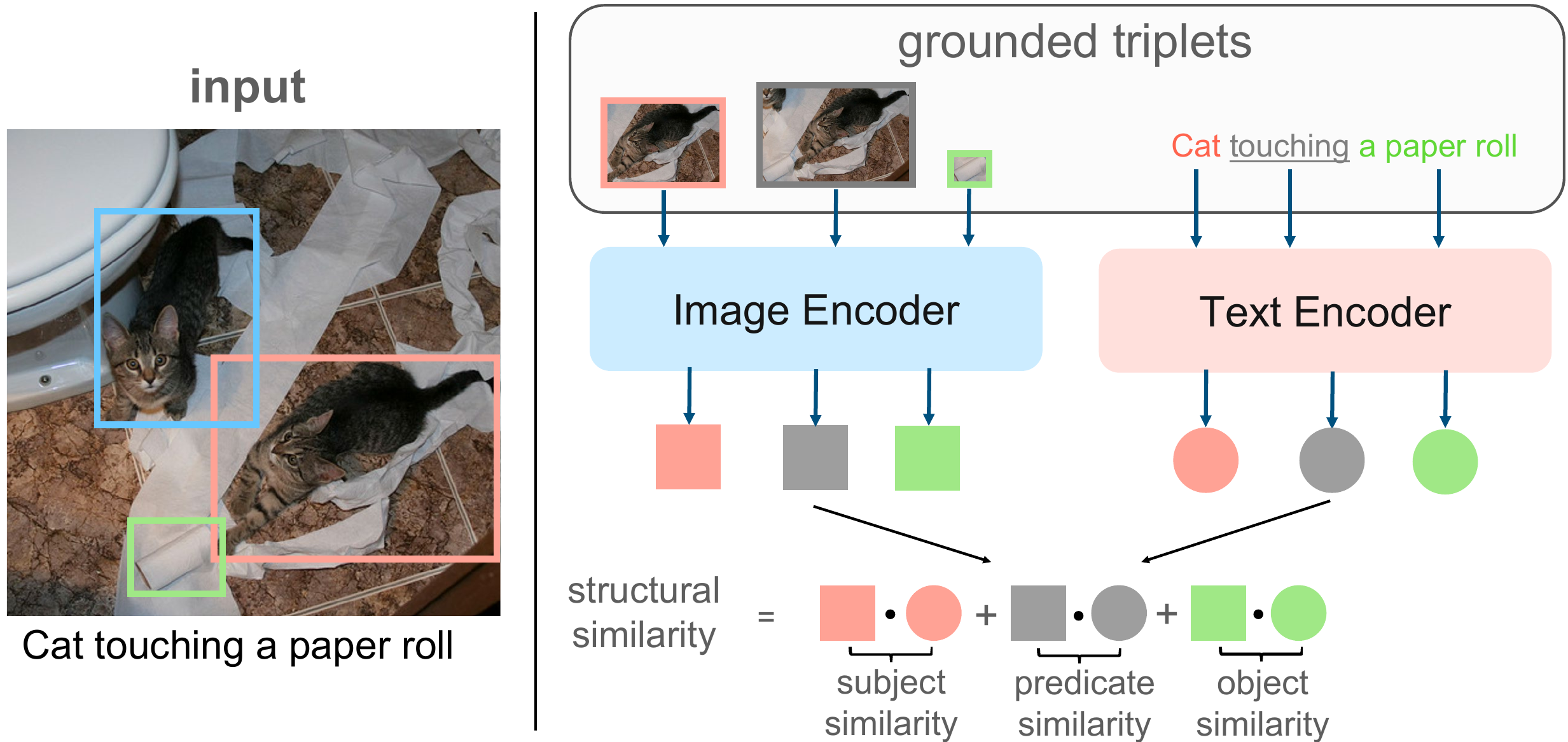}
    \caption{\textbf{Illustration of the triplet-level structural similarity.} Visual and textual triplets are encoded by image encoder and text encoder, respectively. Then the structural similarity is calculated as the sum of cosine similarities between \texttt{subject, predicate,} and \texttt{object}.
    }
    \vspace{-8pt}
    \label{fig:pipeline}
\end{figure}

Our proposed grounding pipeline contains two stages. First, we decouple image and text entities and construct triplets in the format of \((\texttt{subject}, \texttt{predicate}, \texttt{object})\). Second, we calculate triplet-level similarity matrix and propagate it to the instance-level and then obtain the bounding box with the highest similarity score. The primary focus in our matching pipeline is to accurately model the relationship between entities, which is achieved by the the triplet-level structural similarity, as shown in Fig.~\ref{fig:pipeline}. We also provide a novel recipe to equip the VLA models with better compositional understanding ability. 

\begin{figure*}[t]
    \centering
    \includegraphics[width=0.85\linewidth]{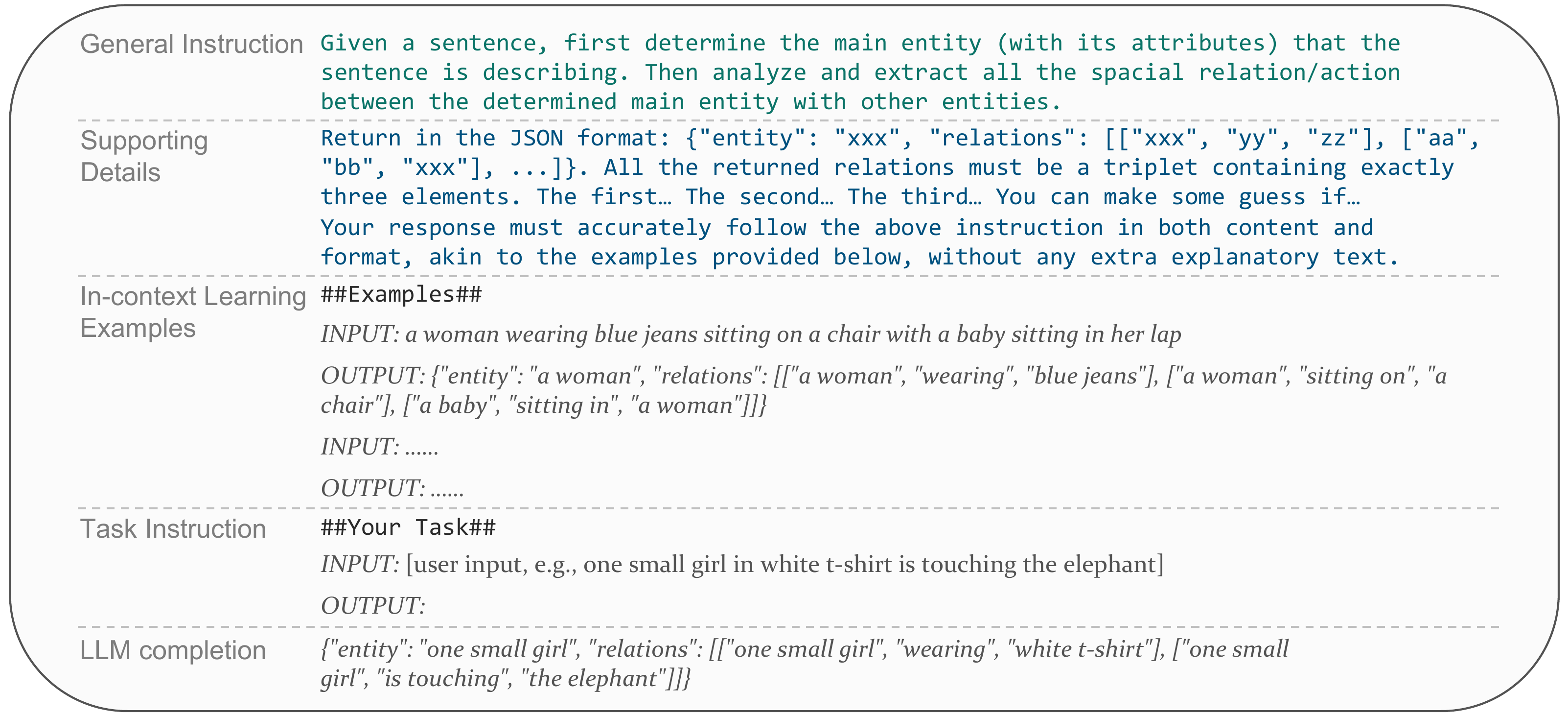}
    \caption{\textbf{Illustration of leveraging ChatGPT's powerful in-context learning capability to parse a caption into triplets.}
    }
    \vspace{-10pt}
    \label{fig:gpt}
\end{figure*}

\subsection{Constructing Triplets}
\label{sec:triplets}
Given a caption, denoted by $\mathcal{C}$, and its associated image, denoted by $\mathcal{I}$, we postulate that both $\mathcal{C}$ and $\mathcal{I}$ comprise sets of entities\footnote{We use ``entities'' to denote objects to differentiate from \texttt{object} in a triplet.}, denoted by $\mathcal{E}_T = \{e_i^T\}_{i=1}^M$ for the text and $\mathcal{E}_I = \{e_i^I\}_{i=1}^N$ for the image, where $M$ and $N$ represent the total number of entities in $\mathcal{C}$ and $\mathcal{I}$, respectively. Interrelation of an entity pair is represented by $r^T(\cdot)$ for the text and $r^I(\cdot)$ for the image. In this stage, our objective is to construct entity-relation triplets for both modalities.

For text, we denote the triplets as:
\begin{equation}
\mathcal{T}_T = \{t_{ij}^T = (e_i^T, r^T(e_i^T, e_j^T), e_j^T) \,|\, 1 \leq k, l \leq M\}.
\end{equation}

For image, we denote the triplets as:
\begin{equation}
\mathcal{T}_I = \{t_{kl}^I = (e_k^I, r^I(e_k^I, e_l^I), e_l^I) \,|\, 1 \leq i, j \leq N\}.
\end{equation}

The cardinality of the above two sets are defined as $M'$ for text and $N'$ for image, respectively.

\noindent\textbf{Textual triplets construction.}
Large language models have exhibited a powerful capacity for a range of downstream tasks. 
Here, we leverage its powerful in-context learning capability to parse a caption $\mathcal{C}$ into triplets $\mathcal{T}_T$.
Specifically, We design a prompt to instruct the ChatGPT to parse the caption text $\mathcal{C}$. Fig.~\ref{fig:gpt} provides an overview on how we design prompt for RefCOCO/+/g dataset, and further details are elaborated as follows. Note that the prompts can vary depending on datasets to accommodate different distributions of the data. 

As shown in Fig.~\ref{fig:gpt}, the prompt can be divided into four parts: \emph{(i) general instruction}; \emph{(ii) supporting details}; \emph{(iii) in-context learning examples}, and \emph{(iv) task instruction}, followed by \emph{LLM completion}, which yields the output of the LLM in the specified format and content. In part (i), we define a clear and general instruction for specific task. Then, we elaborate supporting details in part (ii), including the expected output format, essential elements, and preferences for what should or should not be included, etc. In part (iii), we curate several in-context learning examples to guide the LLM, which is immediately followed by part (iv), where we append the input caption $\mathcal{T}$. Finally, we feed the above input into LLM, and then decoupled textual triplets will be generated through the LLM completion. We also do a simple format check after the completion.

\begin{figure*}[t]
    \centering
    \includegraphics[width=0.75\linewidth]{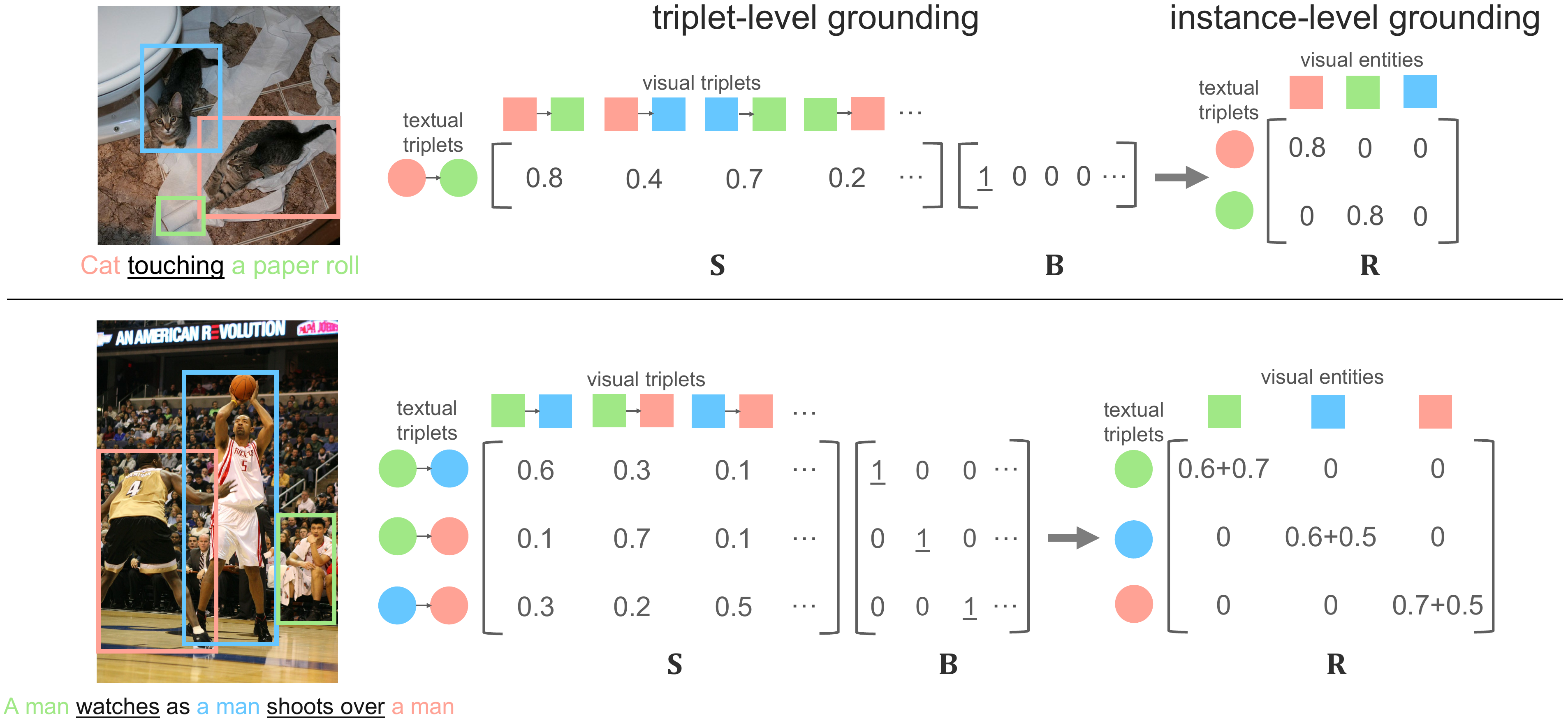}
    \caption{\textbf{Illustration of propagating the similarity scores from grounded triplets to the instance level.}  Via the aggregation of the similarity scores from multiple grounded triplets, it helps find the instance-level correspondences more accurately. For instance, in the lower part, the referring expression \textcolor{lightblue}{a man} and the blue bounding box appear in two different triplets, acting as the \texttt{subject} and \texttt{object}, respectively. Such structural similarity provide more useful cues to improve the instance-level grounding. (Best viewed in color.)}
    \vspace{-10pt}
    \label{fig:grounding}
\end{figure*}

\noindent\textbf{Visual triplets construction.}
In an image, entities are represented by bounding boxes, each enclosing an individual object. These boxes might be predefined by the dataset or extracted using a pre-trained object detector. Without prior knowledge about how these entities are related, we assume potential interactions can happen between every pair of entities. Therefore, we generate visual pairs using a Cartesian product, which includes all possible combinations of entities. A notable case is when a pair consists of the same entity (box) twice. This represents a self-relation, suggesting the entity's own attributes, such as color (\eg, \texttt{red}) or self-actions (\eg, \texttt{walking}). Then we use the union area of two entity boxes to represent $r^I(e_i^I, e_j^I)$, the interrelation between entities.

Here, we derive triplets set $\mathcal{T}_T$ and $\mathcal{T}_I$ from caption and image, respectively. Before moving on, we filter out redundant triplets $t_{ij}^I \in \mathcal{T}_I$ based on heuristic rules similar to ReCLIP \cite{reclip}. Specifically, given a textual triplet $t_{ij}^T$, where its \texttt{predicate} contains keywords that reflect some spacial relationships, \eg, \texttt{to the left of}. In such a case, we filter out visual box pairs where the central point of the former box (\ie, \texttt{subject}) is in the right of the latter one (\ie, \texttt{object}). This approach is much simpler  than building complicated spatial semantic trees like in ReCLIP, yet it effectively adds spatial context and improves performance.

\subsection{Grounding Based on the Structural Similarity}
With the triplets derived from the image and caption, we can use them to solve the visual grounding problem, which allows us to leverage the structural similarity between both modalities to more accurately link the textual descriptions of their corresponding image regions.

We consider two grounding directions: \texttt{text} $\rightarrow$ \texttt{image} and \texttt{image} $\rightarrow$ \texttt{text}, each serving unique task requirements. The \texttt{text} $\rightarrow$ \texttt{image} grounding is applied when identifying specific image regions based on a textual description (\eg, RefCOCO/+/g \cite{refcoco1, refcoco2} dataset). Conversely, \texttt{image} $\rightarrow$ \texttt{text} grounding involves locating relevant textual descriptions for a given image region (\eg, Who's Waldo \cite{whos-waldo} dataset). Given their symmetrical nature, this section will primarily focus on the \texttt{text} $\rightarrow$ \texttt{image} grounding scenario.

\noindent\textbf{Triplet-level grounding.} 
Given a text triplet $t_{ij}^T = (e_i^T, r^T(e_i^T, e_j^T), e_j^T)$, we separately feed $e_i^T$, $r^T(e_i^T, e_j^T)$, and $e_j^T$ into a VLA text encoder to obtain three text embeddings, denoted as $(\vt_{i}, \vt_{i,j}, \vt_{j})$. Similarly, for a image triplet $t_{kl}^I = (e_k^I, r^I(e_k^I, e_l^I), e_l^I)$, we derive three image embeddings $(\vv_{k}, \vv_{k,l}, \vv_{l})$ using the image encoder. The similarity between these two triplets is then given by:
\begin{equation}
\label{eq:sim}
\vS(t_{ij}^T, t_{kl}^I) = \text{cos}(\vt_{i}, \vv_{k}) + \text{cos}(\vt_{i,j}, \vv_{k,l}) + \text{cos}(\vt_{j}, \vv_{l}),
\end{equation}
where $\text{cos}(\cdot,\cdot)$ denotes the cosine similarity function. $\vS\in \mathbb{R}^{M' \times N'}$ is the similarity matrix between all text triplets with all image triplets.
We subsequently get a binary indicator matrix $\vB\in \{0, 1\}^{M' \times N'}$ by:

\begin{equation}
\label{eq:binary}
\vB(t_{ij}^T, t_{kl}^I) = \begin{cases} 
1 & \text{if } k, l = \arg\max_{m,n}(\vS(t_{ij}^T, t_{mn}^I)).\\
0 & \text{otherwise}.\notag
\end{cases}
\end{equation}
Here, for each text triplet $t_{ij}^T$, the binary indicator matrix $\mathbf{B}$  assigns the value of 1 to the most similar image triplet $t_{kl}^I$ and 0 to all others.

\noindent\textbf{Instance-level grounding.} 
Another substantial challenge in the visual grounding problem is that both the \texttt{subject} and \texttt{object} in a triplet may have multiple interactions with other entities. To this end, we design a novel method to propagate the triplet-level grounding results to the instances.

Specifically, based on the triplet-level grounding results, we can compute the instance-level structure-aware similarity matrix $\mathbf{R}$ as follows:
\begin{align}
\label{eq:rk_to_bi}
\vspace{-0.1cm}
     \mathbf{R}(e_i^T, e_k^I) = &\sum_{j, l}\vB(t_{ij}^T, t_{kl}^I) \vS(t_{ij}^T, t_{kl}^I)
     + \notag \\
     & \sum_{j, l}\vB(t_{ji}^T,t_{lk}^I) \vS(t_{ji}^T,t_{lk}^I).
\vspace{-0.1cm}
\end{align}

The two terms in Eq.~\ref{eq:rk_to_bi} consider both the cases where $e_i^T$ and $e_k^I$ appear as the \texttt{subject} and \texttt{object} in different triplets, as shown in lower part of Fig.~\ref{fig:grounding}. By aggregating the similarity scores from multiple grounded triplets, it helps find the instance-level correspondences more accurately.

Finally, for each text entity, we compute the most relevant image entity as follows:
\begin{equation}
\label{eq:instance_selection}
    \hat{e}_k^I = \text{argmax}_m \mathbf{R}(e_i^T, e_m^I),
\end{equation}
where $\hat{e}_i^I$ denotes the corresponding visual entity for $e_i^T$. Notably, our approach can easily extend to the one-to-many grounding scenario if we implement a threshold-based selection in place of the \texttt{argmax} function in Eq.~\ref{eq:binary} and Eq.~\ref{eq:instance_selection}. We limit our discussion to one-to-one grounding for clearer understanding.

\begin{table*}[hbt]
\centering
\begin{tabular}{lcccccccc}
\Xhline{2\arrayrulewidth}
                            & \multicolumn{2}{c}{RefCOCOg}              & \multicolumn{3}{c}{RefCOCO+}                               & \multicolumn{3}{c}{RefCOCO}                    \\
\textbf{Model}              & \textbf{Val} & \textbf{Test}              & \textbf{Val} & \textbf{TestA} & \textbf{TestB}             & \textbf{Val} & \textbf{TestA} & \textbf{TestB} \\ \hline
Random                      & 18.12        & \multicolumn{1}{c|}{19.10} & 16.29        & 13.57          & \multicolumn{1}{c|}{19.60} & 15.73        & 13.51          & 19.20          \\ \hline
Supervised SOTA \cite{uninext} & 88.73        & \multicolumn{1}{c|}{89.37} & 85.24        & 89.63          & \multicolumn{1}{c|}{79.79} & 92.64        & 94.33          & 91.46          \\ \hline
CPT-Blk w/ VinVL \cite{cpt}            & 32.10        & \multicolumn{1}{c|}{32.30} & 25.40        & 25.00          & \multicolumn{1}{c|}{27.00} & 26.90        & 27.50          & 27.40          \\
CPT-Seg w/ VinVL \cite{cpt}           & 36.70        & \multicolumn{1}{c|}{36.50} & 31.90        & 35.20          & \multicolumn{1}{c|}{28.80} & 32.20        & 36.10          & 30.30          \\ \hline
\textbf{CLIP (ViT-B/32)}    &              & \multicolumn{1}{c|}{}      &              &                & \multicolumn{1}{c|}{}      &              &                &                \\
CPT-adapted \cite{reclip}                & 21.77        & \multicolumn{1}{c|}{22.78} & 23.46        & 21.73          & \multicolumn{1}{c|}{26.32} & 23.79        & 22.87          & 26.03          \\
GradCAM \cite{gradcam}                    & 49.51        & \multicolumn{1}{c|}{48.53} & 44.64        & 50.73          & \multicolumn{1}{c|}{39.01} & 42.29        & 49.04          & 36.68          \\
ReCLIP \cite{reclip}                     & 56.96        & \multicolumn{1}{c|}{56.15} & 45.34        & 48.45          & \multicolumn{1}{c|}{42.71} & 45.77        & 46.99          & 45.24          \\
\textbf{Ours}                        & 57.60        & \multicolumn{1}{c|}{56.64} & 45.64        & 47.59          & \multicolumn{1}{c|}{42.79} & 48.24        & 48.40          & 49.15          \\
\textbf{Ours+VR-CLIP}                & 59.87        & \multicolumn{1}{c|}{59.90} & \textbf{55.52}        & \textbf{62.56}          & \multicolumn{1}{c|}{45.69} & \textbf{60.62}        & \textbf{66.52}          & \textbf{54.86}          \\ \hline
\textbf{FLAVA}              &              & \multicolumn{1}{c|}{}      &              &                & \multicolumn{1}{c|}{}      &              &                &                \\
\textbf{Ours}                        & \underline{60.95}        & \multicolumn{1}{c|}{\underline{59.99}} & 48.89        & 50.02          & \multicolumn{1}{c|}{\underline{46.86}}   & 49.37         & 47.76         & 51.68          \\
\textbf{Ours+VR-FLAVA}               & \textbf{61.25}             & \multicolumn{1}{c|}{\textbf{60.86}}      & \underline{50.79}             & \underline{53.35}               & \multicolumn{1}{c|}{\textbf{47.62}}      & \underline{52.46}             & \underline{52.66}               & \underline{52.92}               \\ \Xhline{2\arrayrulewidth}
\end{tabular}
\caption{\textbf{Accuracy on the RefCOCOg, RefCOCO+ and RefCOCO datasets.} \texttt{Ours} represents leveraging our triplet-to-instance pipeline for grounding. \texttt{Ours+VR-CLIP/VR-FLAVA} further replaces the original VLA model with our relationship-enhanced model. Except for the supervised method, the best results are highlighted in \textbf{bold}, and second-best results are \underline{underlined}.}
\label{tab:main-retults}
\end{table*}

\subsection{Enhanced Relational Understanding}
In Equation~\ref{eq:sim}, the interaction between two entities is represented by the term $\text{cos}(\vt_{i,j}, \vv_{k,l})$. This term is crucial as it attempts to quantify the relationship between entities through cosine similarity, under the assumption that VLA models can adequately grasp these relationships. Nevertheless, as indicated by other studies \cite{reclip, bows}, this assumption often falls short in practice. To address this issue, we fine-tuned VLA models using a combination of datasets rich in relational knowledge. These datasets include HICO-det \cite{hico}, SWiG \cite{swig}, and Visual Genome (VG) \cite{vg}. Notably, in the case of the VG dataset, we excluded all the images from COCO to maintain the integrity of the zero-shot protocol, aligning with our experiments based on RefCOCO/+/g.

The datasets mentioned provide annotation bounding boxes for objects together with their textual descriptions and relationships with other objects. So we can easily follow what we have done in the triplet-level grounding stage to create visual-textual triplets, and then utilize a contrastive learning loss on these triplets. To clarify, we use the same notation in Eq.~\ref{eq:sim} to calculate the similarity between two triplets. 
Assume $t_{ij}^T$ and $t_{kl}^I$ are two corresponding triplets, we define the contrastive loss as follows:

\begin{align}
    \mathcal{L} =  \sum_{(t_{ij}^T, t_{kl}^I)} & \left[  \log \left( \frac{\vS(t_{ij}^T, t_{kl}^I)}{\sum_{m,n} \vS(t_{ij}^T, t_{mn}^I)} \right) \right. \notag \\
     & \left. + \log \left( \frac{\vS(t_{ij}^T, t_{kl}^I)}{\sum_{m,n} \vS(t_{mn}^T, t_{kl}^I)} \right) \right].
\end{align}
Through this refined approach, we aim to enhance the VLA models' ability to understand and accurately score the relationship between entities, thereby 
enhancing the zero-shot grounding capability.

Compared with the existing rule-based negative prompts construction \cite{dac, bows, structured-vl}, this design has two unique advantages. 
First, by decomposing a single image into multiple triplets, we can obtain more training instances and improve the diversity of the training data than simply using the entire image for fine-tuning.
Furthermore, 
the isolation of entities removes the distraction of other content in the image, providing more useful supervision for the model fine-tuning.

\section{Experiments}

\subsection{Setup}
\noindent\textbf{RefCOCO/RefCOCO+\cite{refcoco1}/RefCOCOg\cite{refcoco2}} are collected from MS-COCO \cite{coco}. RefCOCO includes 19,994 images with 142,210 referring expressions. RefCOCO+ has 19,992 images and 141,564 expressions. RefCOCOg contains 26,771 images with 104,560 expressions. In RefCOCO and RefCOCO+, expressions are shorter, averaging 1.6 nouns and 3.6 words. In RefCOCOg, expressions are longer, averaging 2.8 nouns and 8.4 words.

\noindent\textbf{Who's Waldo \cite{whos-waldo}} introduces a person-centric visual grounding task, where all names in the captions are masked, forcing models to link boxes and the masked [NAME] tokens through attributes and interactions between visual entities. The captions are long and contain complex scene descriptions. We use its test split for evaluation, which contains 6741 images. Each caption contains about 30 words.

\noindent\textbf{Evaluation metrics.} On RefCOCO/+/g, we follow previous work \cite{reclip} to use accuracy as the grounding results, \ie, if the IoU (Intersection over Union) value between the predicted box and ground truth region is larger than 0.5, it is a correct prediction. On Who's Waldo, following previous work \cite{whos-waldo}, given the grounding results of person in textual descriptions to bounding boxes in the image, we report the accuracy against ground-truth links on the test set.

\subsection{Implementation Details}
\noindent\textbf{RefCOCO/+/g}
On RefCOCO/+/g datasets, we adopt the test-time augmentation strategy outlined in ReCLIP~\cite{reclip}, where they use both cropping and blurring for isolated visual entities. When blurring the union region, we separately process each box in the box pair, isolating only the area where the box is located, rather than directly using the union area, which helps to minimize distraction from other visual objects. We use the whole caption instead of the decoupled main entities before feeding into the text encoder since it produces better results. All the heuristic rules mentioned in Sec.~\ref{sec:triplets} are also adopted from ReCLIP.

\noindent\textbf{VLA fine-tuning}
We fine-tune the CLIP model based on the code from \cite{structured-vl}, where LoRA rank $r = 4$, batch size is 1024, learning rate is $5e-6$, and epoch is 20. We use the huggingface PEFT \cite{peft} for fine-tuning FLAVA, where we set lora rank $r = 16$ for 10 epochs. Since the SWiG dataset \cite{swig} does not contains triplet annotations, we use ChatGPT to convert the existing annotation into triplets. 

\noindent\textbf{Box generator}
Following ReCLIP, we use bounding boxes generated from MAttNet \cite{mattnet} as the box proposals on RefCOCO/+/g. On the Who's Waldo dataset, we use the box proposals provided in the annotations.

\begin{table}[!t]
\centering

\begin{tabular}{lc}
\Xhline{2\arrayrulewidth}
\textbf{Method}      & \textbf{Test Accuracy} \\ \hline
\textbf{Supervised}                            & \multicolumn{1}{l}{}                    \\
\textit{Who's Waldo} \cite{whos-waldo} & \textbf{63.5}                                    \\ \hline
\multicolumn{2}{l}{\textbf{Pretrained on grounding data}}                                             \\
Gupta et al. \cite{gupta2020contrastive} (COCO)                   & 35.6                                    \\
Gupta et al. \cite{gupta2020contrastive} (Flickr30K)     & 38.2                                    \\
SL-CCRF \cite{liu2020consnet}                               & 46.4                                    \\
MAttNet \cite{mattnet}                              & 24.1                                    \\
UNITER \cite{Uniter}                               & 34.2                                    \\ \hline
\textbf{CLIP}                                             \\
ReCLIP \cite{reclip}                               & 29.4                                    \\ 
\textbf{Ours}                           & 60.8                                    \\
\textbf{Ours+VR-CLIP}                        & \underline{61.3}                                \\ \hline
\textbf{FLAVA}                                             \\
\textbf{Ours}                         & 59.6                                    \\
\textbf{Ours+VR-FLAVA}                        & 59.8                                \\ \Xhline{2\arrayrulewidth}
\end{tabular}

\caption{\textbf{Accuracy on the Who's Waldo dataset.} The best results are highlighted in \textbf{bold}, and second-best results are \underline{underlined}.}
\label{tab:whos-waldo}
\vspace{-0.2cm}
\end{table}

\subsection{Main Results}
\noindent\textbf{RefCOCO/+/g}
We benchmark our approach against various zero-shot visual grounding models, including Colorful Prompt Tuning (CPT) \cite{cpt}, GradCAM \cite{gradcam}, and ReCLIP \cite{reclip}. CPT-adapted is introduced and adapted by \cite{reclip}. ReCLIP represents the latest SOTA in zero-shot REC methods.

As shown in Table~\ref{tab:main-retults}, compared to other models utilizing the same CLIP architecture, our proposed method consistently outperforms all competitors across all splits. Specifically, our model exhibits a performance improvement of up to 19.53\% over ReCLIP, with an average enhancement of 9.74\%. Remarkably, even without fine-tuning the backbone CLIP model, our method can surpass the ReCLIP by up to 3.91\%, with an average of 1.05\%, showing that the structural similarity based on ChatGPT's parsing also contributes to the relational understanding.

In addition, we also extend our methodology to another VLA model, FLAVA~\cite{flava}, to verify the generalizability of our approach. Not surprisingly, when integrated into our matching pipeline, FLAVA demonstrates superior performance compared to the CLIP model. This can be attributed to FLAVA's inherently more robust architecture. After fine-tune the FLAVA, the resulting VR-FLAVA consistently improve the performance across all dataset splits, reinforcing the effectiveness of our method in enhancing the relationship understanding of various VLA models.

\noindent\textbf{Who's Waldo}
We compare our approach with the 
models trained on grounding dataset, which include Gupta et al. \cite{gupta2020contrastive}, SL-CCRF \cite{liu2020consnet}, MAttNet \cite{mattnet} and UNITER \cite{Uniter}. The \emph{Who's Waldo} method serves as a supervised baseline, as reported in its original paper~\cite{whos-waldo}. Additionally, we adapte ReCLIP~\cite{reclip} for our dataset, utilizing their language parser to identify potential referring expressions, followed by grounding using their original approach\footnote{The experiment is conducted using the authors' released code.}.

\begin{table}[t]
    \centering
    \begin{tabular}{c|c|c|c}
    \toprule
    Triplet & VR-CLIP & \textbf{Val} & \textbf{Test} \\
    \midrule
    \xmark & \xmark &  55.35 & 54.33 \\
    \xmark & \cmark & 56.90 & 56.81 \\
    \cmark & \xmark & 57.60 & 56.64 \\
    \cmark & \cmark & \textbf{59.87} & \textbf{59.90}\\
    \bottomrule
    \end{tabular}
    \caption{\textbf{Effectiveness of each component of our grounding pipeline on RefCOCOg.} Triplet means whether we use the triplet-to-instance grounding (instead of scoring-and-ranking). VR-CLIP represents whether we use fine-tuned VR-CLIP instead of the original CLIP model.}
    \label{tab:ablation}
\vspace{-0.2cm}
\end{table}

As shown in Table~\ref{tab:whos-waldo}, 
our approach outperforms all models trained on the grounding dataset, with a notable margin. Among these, SL-CCRF was the closest competitor, yet it falls behind our method by 14.9\%. When compared to the supervised \emph{Who's Waldo} method, our zero-shot setting only shows a 2.2\% reduction in performance. This highlights our method's effectiveness in visual grounding, especially in processing long and complex captions. Notably, performance of ReCLIP is no better than random choice. This is because their language parser fails to handle the real-world complex captions as in Who's Waldo. 
It validates our design choice of using the in-context learning capability of LLMs for better generalization ability.

\begin{figure*}[htb]
    \centering
    \includegraphics[width=0.825\linewidth]{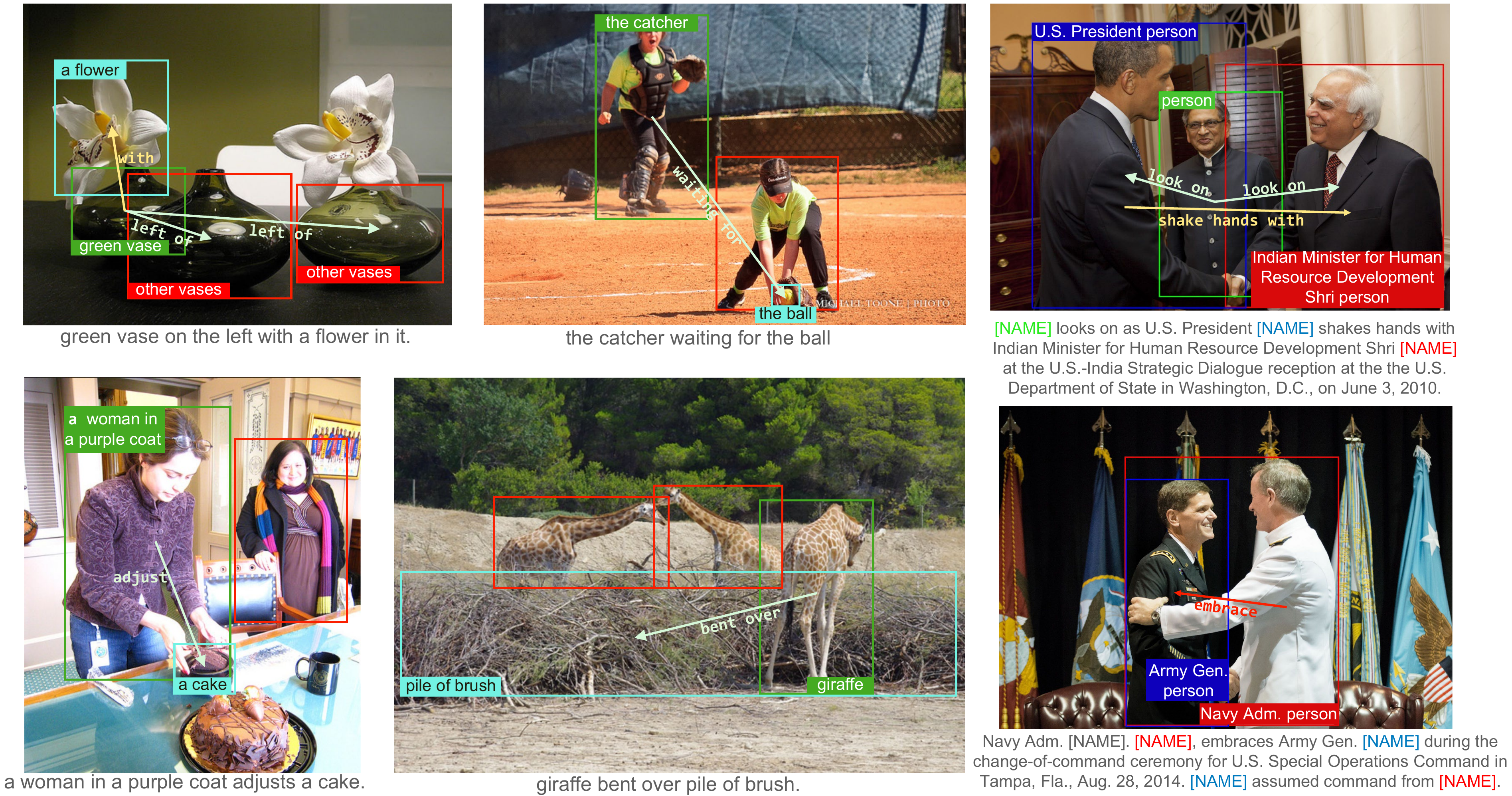}
    \caption{\textbf{Zero-shot visual grounding results.} Left two columns are results from RefCOCO, where our predictions are in green box, distraction objects are in red box. The rightmost column shows results from Who's Waldo, where predicted annotation links are in the same color. Arrows represent relationships between visual objects, and the text on the images are the parsed triplets.}
    \label{fig:vis}
\end{figure*}

We show visualization results in Fig.~\ref{fig:vis}. The left two columns display results from RefCOCOg, while the rightmost column shows results from Who's Waldo. ReCLIP failed in all examples listed. Our approach, which explicitly models relationships (indicated by arrows in the images), provides more helpful information for grounding. Additionally, it is observed that ChatGPT consistently excels in parsing complex captions, such as those in Who's Waldo.

\subsection{Ablation Studies}
In this section, we conduct ablation studies on RefCOCOg. This dataset is particularly suitable for our evaluation because it has longer captions and rich entity interactions, making it an ideal testbed for assessing each component. 

\noindent\textbf{Effectiveness of components in the grounding pipeline.} 
We explore two key variations: \emph{Triplet} and \emph{VR-CLIP}. The \emph{Triplet} variation examines the impact of utilizing triplet-to-instance matching as opposed to a basic scoring-and-ranking approach, i.e., scoring each isolated boxes using CLIP, than select one with the highest similarity. The \emph{VR-CLIP} variant assesses the performance differences between the fine-tuned VR-CLIP and the original CLIP model.

As shown in Table~\ref{tab:ablation}, substituting the CLIP model with the VR-CLIP yields superior results because of the enhanced relational capability. Note that although we do not explicitly use the structural similarity in this context, we are still using the whole caption as the referring expressions, which inherently convey the relationship information. Further analysis reveals that if we replace the scoring-and-ranking with our proposed triplet-to-instance matching pipeline, we can get better results through the relationship modeling. By combining both, we can achieve best results.

\begin{table}[t]
\vspace{-0.2cm}
    \centering
    \begin{tabular}{c|c|c}
    \toprule
      & \textbf{Val} & \textbf{Test} \\
    \midrule
     full & \textbf{59.87}  & \textbf{59.90} \\
     w/o \texttt{subject} & 48.35 & 47.83 \\
     w/o \texttt{object} & 56.92 & 57.05 \\
    w/o \texttt{predicate} & 56.90 & 56.81 \\
    \bottomrule
    \end{tabular}
    \caption{\textbf{Effectiveness of each triplet component on RefCOCOg.} Each removal is done by set the corresponding term in Eq.~\ref{eq:sim} to 0. For w/o \texttt{predicate}, we also turn off the box pair filter to remove any spatial information provided in the caption.}
    \vspace{-10pt}
    \label{tab:wo}

\end{table}

\noindent\textbf{Effectiveness of triplet components.} 
We separately remove the \texttt{subject}, \texttt{object} and \texttt{predicate} terms in Eq.~\ref{eq:sim} to explore their effectiveness in grounding performance. As shown in Table~\ref{tab:wo}, the absence of \texttt{subject} or \texttt{object} leads to the loss of most noun information and its attributes, resulting in a clear accuracy drop. This impact is particularly significant for \texttt{subject} removal, since most main entities in RefCOCOg are represented as \texttt{subject}. Without \texttt{predicate},  meaning no interrelation between entities is considered, the accuracy degrades by about 3 points. It emphasizes the importance of our structural similarity in modeling the entity relationships.

\noindent\textbf{Effectiveness of triplet- to instance-level grounding .} 
We separately remove the first and second terms in Eq.~\ref{eq:rk_to_bi} to validate our design of triplet- to instance-level grounding. The results are reported in Table~\ref{tab:eq}, which shows that utilizing both pieces of information yields the best performance.

\subsection{Limitations}

While our method enhances the visual relationship understanding of VLA models, it sacrifices the model's zero-shot capabilities as generalist models and downgrades them to specialist ones. For example, after fine-tuning, CLIP's zero-shot image classification accuracy decreases from 0.63 to 0.50 and the R@5 for image retrieval on COCO decreases from 0.57 to 0.54. A promising future direction is to apply large-scale unlabeled image-caption pairs during VLA fine-tuning, and probably we can generate region-text triplets using ChatGPT and SAM (Segment Anything Model)~\cite{kirillov2023segment} as pseudo ground-truths.

\begin{table}[t]
\vspace{-0.2cm}
    \centering
    \begin{tabular}{c|c|c}
    \toprule
      & \textbf{Val} & \textbf{Test} \\
    \midrule
     full & \textbf{59.87}  & \textbf{59.90} \\
     w/o 1st term & 59.54 & 59.37 \\
     w/o 2nd term & 59.11 & 59.26 \\
    \bottomrule
    \end{tabular}
    \caption{\textbf{Effectiveness of different terms in Eq.~\ref{eq:rk_to_bi} for triplet- to instance-level grounding on RefCOCOg.}}
    \vspace{-10pt}
    \label{tab:eq}
\end{table}

\section{Conclusion}
In this paper, we proposed a novel zero-shot referring expression comprehension model by resorting to powerful capabilities of foundation models (\eg, ChatGPT and CLIP) to explicitly model the structural similarity between entities and then find their correspondences by propagating the similarity from triplets to instances.
We also introduced a novel recipe to improve a VLA model's visual relationship understanding by training from a collection of curated data sources. Experimental results on RefCOCO/+/g and Who's Waldo validate the effectiveness of our approach.

\appendix
\section{Implementation Details}
\noindent\textbf{Special Cases for Textual Triplets.}
Not all caption can be perfectly decoupled into textual triplets. Often, such a caption is just a single noun or lacks an explicit subject. For example, the caption \texttt{red apple} would be decoupled into (\texttt{"red apple"}, \texttt{""}, \texttt{""}), and \texttt{person walking} into (\texttt{"person"}, \texttt{"walking"}, \texttt{""}). In these instances, we fill the blank spaces (\ie, missing items in the triplet) with the \texttt{subject} string, resulting in (\texttt{"red apple"}, \texttt{"red apple"}, \texttt{"red apple"}) and (\texttt{"person"}, \texttt{"walking"}, \texttt{"person"}), respectively, for the previous examples. This approach ensures our grounding pipeline can manage such simplified cases. For instance, (\texttt{"red apple"}, \texttt{"red apple"}, \texttt{"red apple"}) will be matched three times with the visual entity of a red apple, which means that it degenerates to the naive score-and-ranking strategy. 

Additionally, before feeding the \texttt{predicate} into the text encoder, on the RefCOCO/+/g dataset, we form a complete sentence by concatenating the \texttt{subject}, \texttt{predicate} and \texttt{object}, \eg, \texttt{"vase on top of table"} instead of \texttt{"on top of"}.
On the Who's Waldo dataset, we add \texttt{a person} before and after the \texttt{predicate}, \eg, \texttt{"a person looking at a person"} instead of \texttt{"looking at"}.
This is because, in most cases, a single \texttt{predicate} like \texttt{"on top of"} is semantically meaningless. Instead, a complete phrase like \texttt{"vase on top of table"} offers more contextual information.

\begin{figure}[t]
    \centering
    \includegraphics[width=0.9\linewidth]{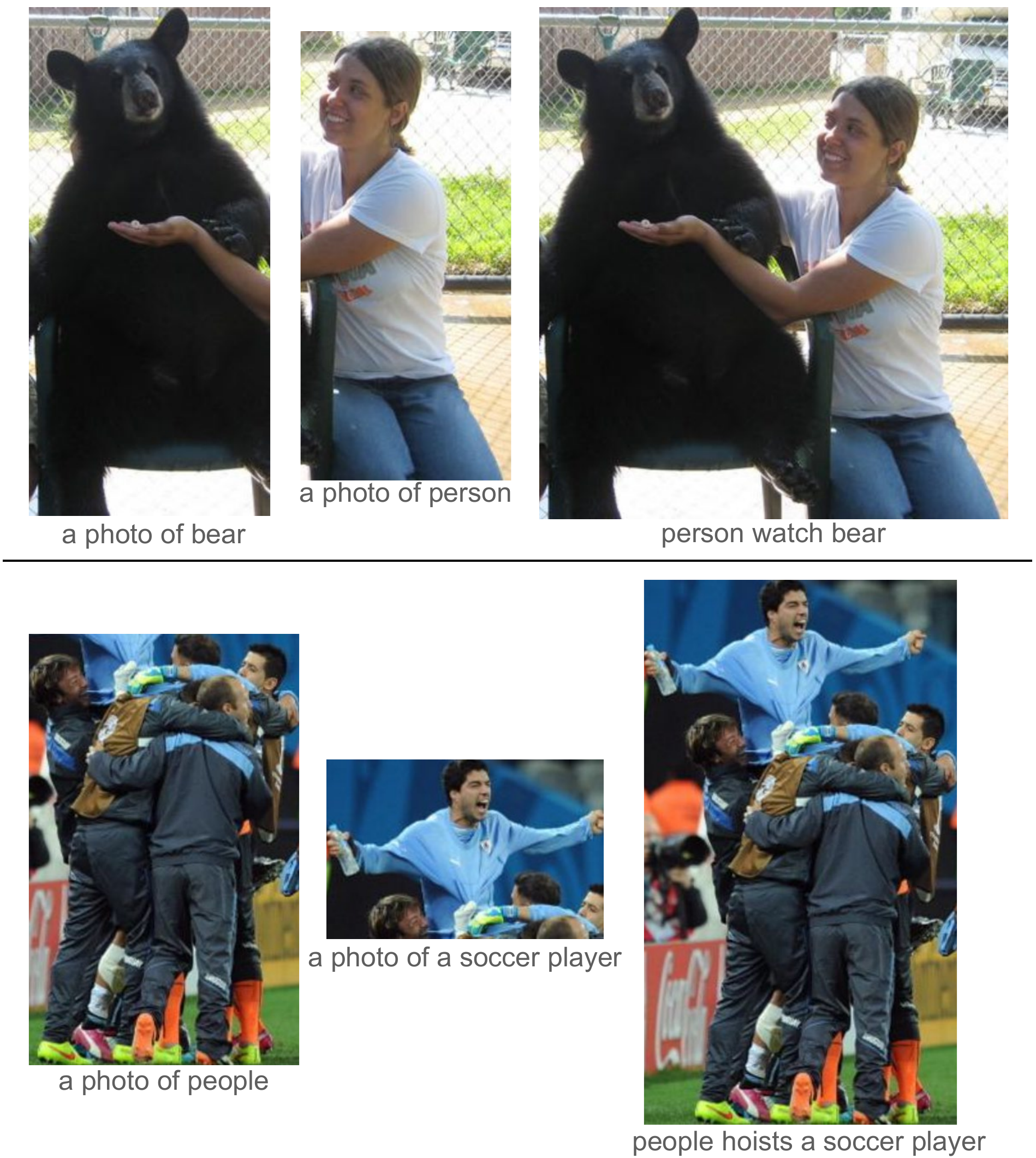}
    \caption{\textbf{Two examples in the dataset for VLA fine-tuning.}}
    \label{fig:vla}
\end{figure}

\noindent\textbf{Dataset for VLA Fine-tuning.}
Our dataset for VLA fine-tuning is obtained from HICO-det~\cite{hico}, SWiG~\cite{swig} and Visual Genome (without COCO images)~\cite{vg}. Each datapoint in this dataset consists of multiple image-text triplet pairs with the same text triplet, with two examples illustrated in Fig.~\ref{fig:vla}. Notably, ``multiple'' is because we group image triplets corresponding to the same textual triplet into a single datapoint. Consequently, a single datapoint may comprise several distinct image triplets paired with the same textual triplet. For training purposes, we randomly select one image triplet from each category per epoch. This strategy is adopted to avoid scenarios in a single batch where an image triplet is forced to simultaneously pull in and push away from the same textual triplet due to the contrastive learning objective.

\section{Additional Experiment Results}
This section provides additional experiment results tested on RefCOCO/+/g. Table.~\ref{tab:supp-results} shows the full results with different box proposal variants, \ie, using a bounding box size prior (filter our objects smaller than 5\% of the image), and use the groundtruth bounding boxes as box proposals.

\begin{table*}[hbt]
\centering
\resizebox{0.9\linewidth}{!}{
\begin{tabular}{lcccccccc}
\Xhline{3\arrayrulewidth}
                            & \multicolumn{2}{c}{RefCOCOg}              & \multicolumn{3}{c}{RefCOCO+}                               & \multicolumn{3}{c}{RefCOCO}                    \\
\textbf{Model}              & \textbf{Val} & \textbf{Test}              & \textbf{Val} & \textbf{TestA} & \textbf{TestB}             & \textbf{Val} & \textbf{TestA} & \textbf{TestB} \\ \hline
Random                      & 18.12        & \multicolumn{1}{c|}{19.10} & 16.29        & 13.57          & \multicolumn{1}{c|}{19.60} & 15.73        & 13.51          & 19.20          \\ 
Random (w/ groundtruth box proposal)                     & 20.18        & \multicolumn{1}{c|}{20.34} & 16.73        & 12.57          & \multicolumn{1}{c|}{22.13} & 16.37        & 12.45          & 21.32          \\ \hline
Supervised SOTA \cite{uninext} & 88.73        & \multicolumn{1}{c|}{89.37} & 85.24        & 89.63          & \multicolumn{1}{c|}{79.79} & 92.64        & 94.33          & 91.46          \\ \hline
CPT-Blk w/ VinVL \cite{cpt}            & 32.10        & \multicolumn{1}{c|}{32.30} & 25.40        & 25.00          & \multicolumn{1}{c|}{27.00} & 26.90        & 27.50          & 27.40          \\
CPT-Seg w/ VinVL \cite{cpt}           & 36.70        & \multicolumn{1}{c|}{36.50} & 31.90        & 35.20          & \multicolumn{1}{c|}{28.80} & 32.20        & 36.10          & 30.30          \\ \Xhline{2\arrayrulewidth}
\textbf{CLIP}    &              & \multicolumn{1}{c|}{}      &              &                & \multicolumn{1}{c|}{}      &              &                &                \\
CPT-adapted \cite{reclip}                & 21.77        & \multicolumn{1}{c|}{22.78} & 23.46        & 21.73          & \multicolumn{1}{c|}{26.32} & 23.79        & 22.87          & 26.03          \\
GradCAM \cite{gradcam}                    & 49.51        & \multicolumn{1}{c|}{48.53} & 44.64        & 50.73          & \multicolumn{1}{c|}{39.01} & 42.29        & 49.04          & 36.68          \\
ReCLIP \cite{reclip}                     & 56.96        & \multicolumn{1}{c|}{56.15} & 45.34        & 48.45          & \multicolumn{1}{c|}{42.71} & 45.77        & 46.99          & 45.24          \\
\textbf{Ours}                        & 57.60        & \multicolumn{1}{c|}{56.64} & 45.64        & 47.59          & \multicolumn{1}{c|}{42.79} & 48.24        & 48.40          & 49.15          \\
\textbf{Ours+VR-CLIP}                & \textbf{59.87}        & \multicolumn{1}{c|}{\textbf{59.90}} & \textbf{55.52}        & \textbf{62.56}          & \multicolumn{1}{c|}{\textbf{45.69}} & \textbf{60.62}        & \textbf{66.52}          & \textbf{54.86}          \\ \hline

\textbf{CLIP (w/ box size prior)}    &              & \multicolumn{1}{c|}{}      &              &                & \multicolumn{1}{c|}{}      &              &                &                \\
CPT-adapted \cite{reclip}                & 28.98        & \multicolumn{1}{c|}{30.14} & 26.64        &  25.13          & \multicolumn{1}{c|}{27.27} & 26.08        &  25.38          & 28.03          \\
GradCAM \cite{gradcam}                    &  52.29        & \multicolumn{1}{c|}{51.28} & 49.41        & 59.66          & \multicolumn{1}{c|}{38.62} &  44.65        &  53.49          &  36.19          \\
ReCLIP \cite{reclip}                     &  \textbf{60.85}        & \multicolumn{1}{c|}{\textbf{61.05}} & 55.07        & 60.47          & \multicolumn{1}{c|}{47.41} & 54.04        & 58.60          & 49.54          \\
\textbf{Ours}                        & 58.52        & \multicolumn{1}{c|}{57.95} & 52.38        & 57.65          & \multicolumn{1}{c|}{45.65} & 56.10        & 58.97          & 52.23          \\
\textbf{Ours+VR-CLIP}                & 58.95        & \multicolumn{1}{c|}{59.55} & \textbf{58.65}        & \textbf{68.32}          & \multicolumn{1}{c|}{\textbf{47.42}} & \textbf{62.92}        & \textbf{69.90}          & \textbf{55.19}          \\ \hline
\textbf{CLIP (w/ groundtruth box proposal)}    &              & \multicolumn{1}{c|}{}      &              &                & \multicolumn{1}{c|}{}      &              &                &                \\
CPT-adapted \cite{reclip}                & 24.16        & \multicolumn{1}{c|}{24.70} & 25.07        & 22.28          & \multicolumn{1}{c|}{28.68} & 25.12        & 23.39          & 28.42          \\
GradCAM \cite{gradcam}                    &  54.00        & \multicolumn{1}{c|}{54.01} & 48.00        & 52.13          & \multicolumn{1}{c|}{43.85} &  45.41        & 50.13          & 41.47          \\
ReCLIP \cite{reclip}                     & \textbf{65.48}        & \multicolumn{1}{c|}{64.38} & 49.20        & 50.23          & \multicolumn{1}{c|}{48.58} & 49.69        & 48.08          & 52.50          \\
\textbf{Ours}                        & 64.99        & \multicolumn{1}{c|}{64.03} & 49.75        & 50.18          & \multicolumn{1}{c|}{49.77} & 52.82        &  49.90         & 57.29          \\
\textbf{Ours+VR-CLIP}                & 65.11        & \multicolumn{1}{c|}{\textbf{66.00}} & \textbf{58.65}        & \textbf{64.78}          & \multicolumn{1}{c|}{\textbf{53.98}} & \textbf{65.60}        & \textbf{68.59}          & \textbf{63.51}          \\

\Xhline{2\arrayrulewidth}
\textbf{FLAVA}              &              & \multicolumn{1}{c|}{}      &              &                & \multicolumn{1}{c|}{}      &              &                &                \\
\textbf{Ours}                        & 60.95        & \multicolumn{1}{c|}{59.99} & 48.89        & 50.02          & \multicolumn{1}{c|}{46.86}   & 49.37         & 47.76         & 51.68          \\
\textbf{Ours+VR-FLAVA}               & \textbf{61.25}             & \multicolumn{1}{c|}{\textbf{60.86}}      & \textbf{50.79}             & \textbf{53.35}               & \multicolumn{1}{c|}{\textbf{47.62}}      & \textbf{52.46}             & \textbf{52.66}               & \textbf{52.92}               \\ \hline
\textbf{FLAVA (w/ box size prior)}              &              & \multicolumn{1}{c|}{}      &              &                & \multicolumn{1}{c|}{}      &              &                &                \\
\textbf{Ours}                        & 60.40        & \multicolumn{1}{c|}{60.73} & 54.82        & 59.73          & \multicolumn{1}{c|}{\textbf{48.25}}   & 57.22         &  59.61        & 55.05          \\
\textbf{Ours+VR-FLAVA}               & \textbf{60.48}             & \multicolumn{1}{c|}{\textbf{61.28}}      & \textbf{55.00}             & \textbf{61.13}               & \multicolumn{1}{c|}{48.17}      & \textbf{57.80}             & \textbf{60.86}               & \textbf{55.33}               \\ \hline
\textbf{FLAVA (w/ groundtruth box proposal)}              &              & \multicolumn{1}{c|}{}      &              &                & \multicolumn{1}{c|}{}      &              &                &                \\
\textbf{Ours}                        & 67.71        & \multicolumn{1}{c|}{66.11} & 52.17        & 51.73          & \multicolumn{1}{c|}{54.33}   & 55.75         & 50.68         & 62.10          \\
\textbf{Ours+VR-FLAVA}               & \textbf{67.97}             & \multicolumn{1}{c|}{\textbf{67.25}}      & \textbf{54.66}             & \textbf{55.78}               & \multicolumn{1}{c|}{\textbf{54.82}}      & \textbf{58.22}             & \textbf{55.83}               & \textbf{62.47}               \\ \Xhline{3\arrayrulewidth}

\end{tabular}
}
\caption{\textbf{Accuracy on the RefCOCOg, RefCOCO+ and RefCOCO datasets.} \texttt{Ours} represents leveraging our triplet-to-instance pipeline for grounding. \texttt{Ours+VR-CLIP/VR-FLAVA} further replaces the original VLA model with our relationship-enhanced model. Results excluding object boxes smaller than 5\% of the image size are denoted as \texttt{w/ box size prior}. Results using groundtruth box proposals are indicated as \texttt{w/ groundtruth box proposal}. For every combination of model and box proposal type, the best results are highlighted in \textbf{bold}.}
\label{tab:supp-results}
\end{table*}

\section{Additional Visualization Results}
\begin{figure*}[t]
    \centering
    \includegraphics[width=0.86\linewidth]{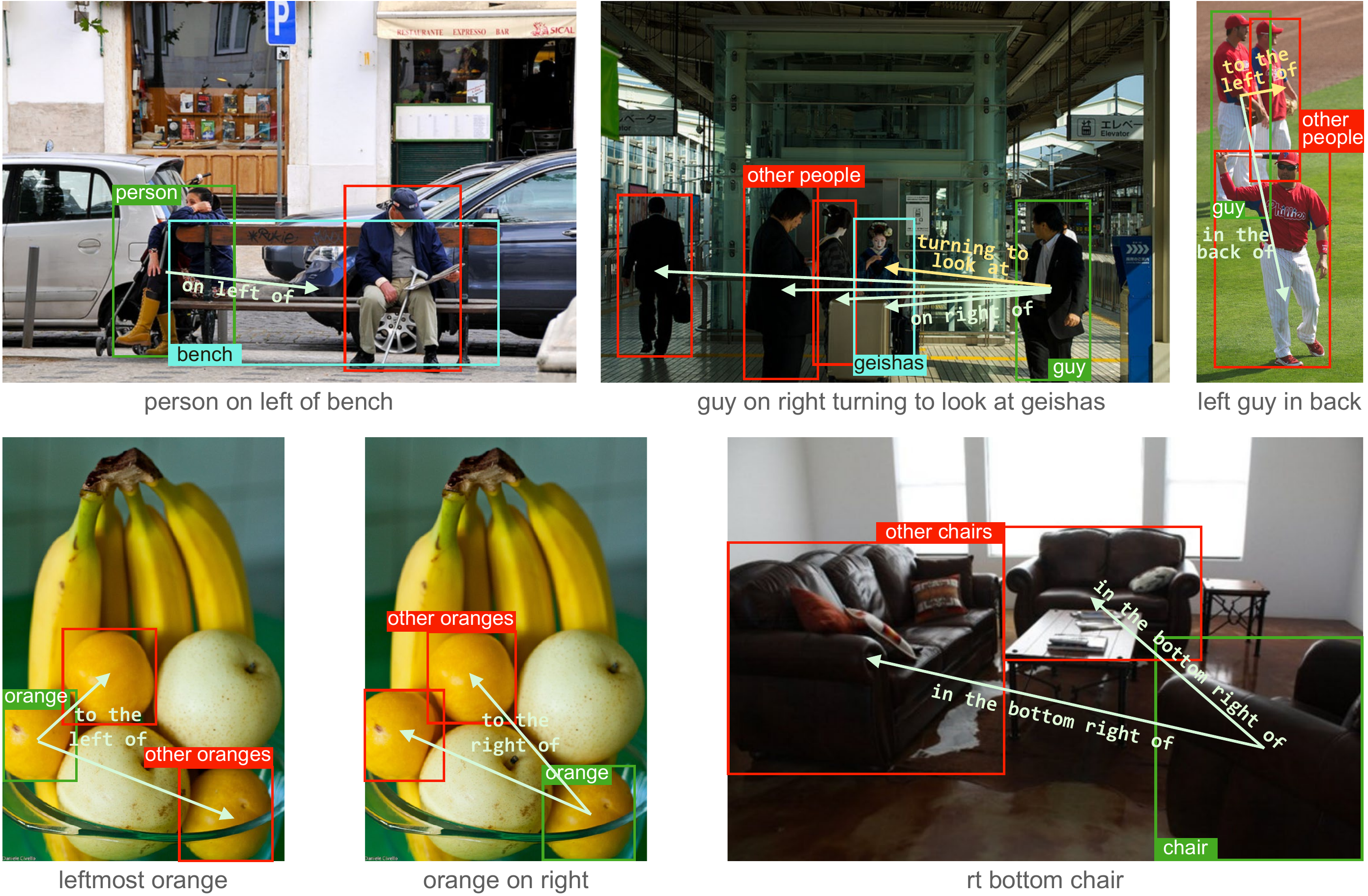}
    \caption{\textbf{Zero-shot visual grounding results on RefCOCO.} Our predictions are in green box, distraction objects are in red box. Arrows represent relationships between visual objects, and the text on the images are the parsed triplets.}
    \label{fig:vis1}
\end{figure*}

\begin{figure*}[!htb]
    \centering
    \includegraphics[width=0.86\linewidth]{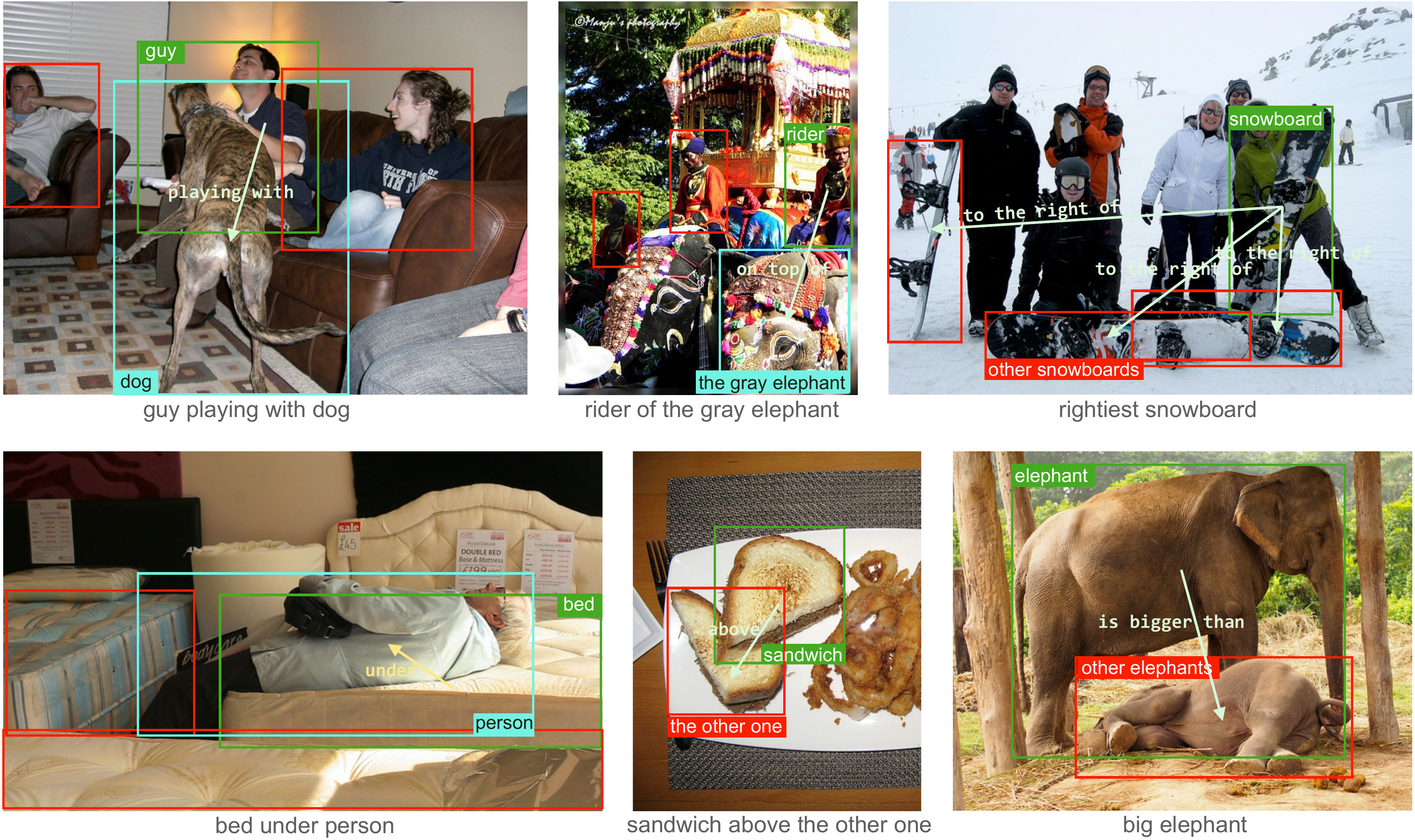}
    \caption{\textbf{Zero-shot visual grounding results on RefCOCO+.} Our predictions are in green box, distraction objects are in red box. Arrows represent relationships between visual objects, and the text on the images are the parsed triplets.}
    \label{fig:vis2}
\end{figure*}

\begin{figure*}[!htb]
    \centering
    \includegraphics[width=0.86\linewidth]{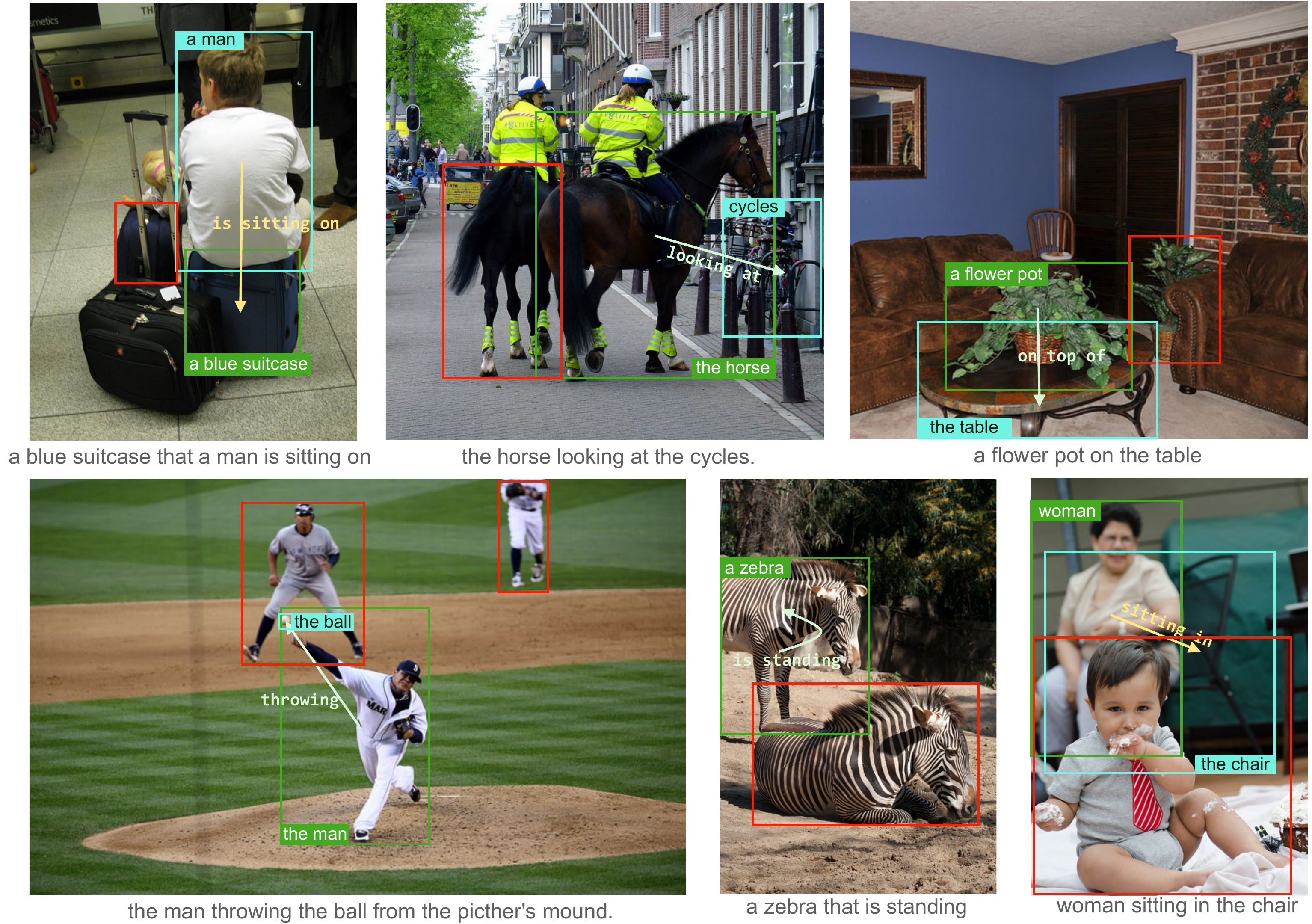}
    \caption{\textbf{Additional zero-shot visual grounding results on RefCOCOg.} Our predictions are in green box, distraction objects are in red box. Arrows represent relationships between visual objects, and the text on the images are the parsed triplets.}
    \label{fig:vis3}
\end{figure*}

\begin{figure*}[!htb]
    \centering
    \includegraphics[width=0.86\linewidth]{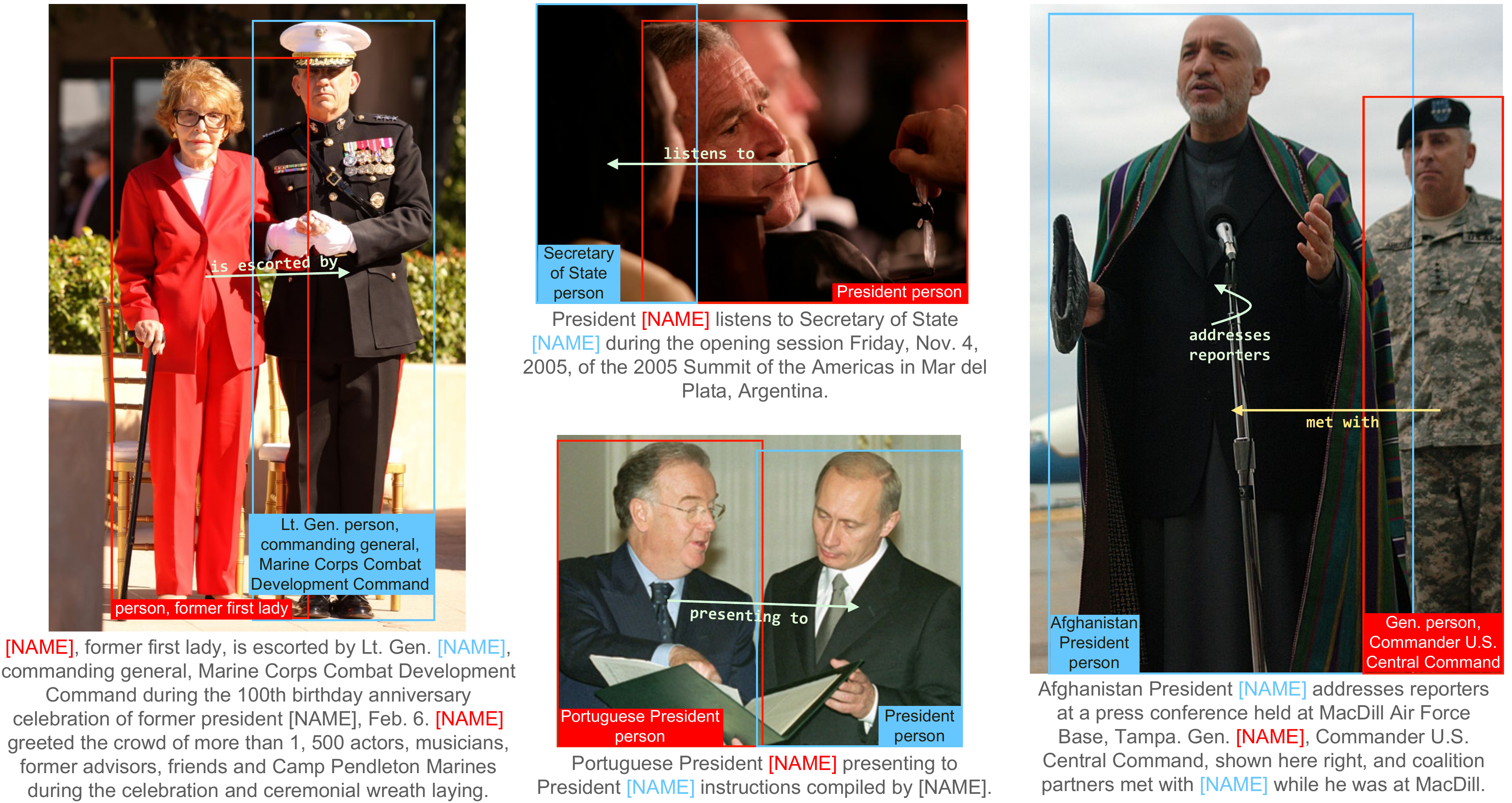}
    \caption{\textbf{Additional zero-shot visual grounding results on Who's Waldo.} Predicted annotation links are in the same color. Arrows represent relationships between visual objects, and the text on the images are the parsed triplets.}
    \label{fig:vis4}
\end{figure*}

In this section, we provide additional visualization results for RefCOCO, RefCOCO+, RefCOCOg, and Who's Waldo, illustrated in Fig.\ref{fig:vis1}, Fig.\ref{fig:vis2}, Fig.\ref{fig:vis3}, and Fig.\ref{fig:vis4}, respectively. Fig.~\ref{fig:vis1}, \ref{fig:vis2}, \ref{fig:vis3} highlight examples where ReCLIP failed but our grounding approach yielded correct results. The textual triplets parsed by ChatGPT are also displayed in the images. Next, we will discuss some selected examples to illustrate the advantages of our approach.

As shown in Fig.~\ref{fig:vis3}, on the RefCOCOg dataset, our approach is able to successfully differentiate multiple instances of the same object category by understanding their relationships with others.
For instance, in the first example in Fig.~\ref{fig:vis3}, the blue suitcase can be accurately grounded among other ones.
For the example of ``a zebra that is standing'', the curved arrow in image represents self-action, where no \texttt{object} is involved, which is a special case of our grounding pipeline.

ChatGPT plays an important role in improving the robustness of our grounding approach. In Fig.~\ref{fig:vis1}, as for ``rt bottom chair'', ChatGPT understands that ``rt'' stands for ``right'', allowing us to accurately generate triplets as depicted in the image. Similarly, in Fig.~\ref{fig:vis2}, it is worth highlighting the example ``rider of the gray elephant''. Here, ChatGPT made some reasonable deduction that rider is ``on top of'' the elephant. With longer captions, as shown in Figure \ref{fig:vis2}, ChatGPT can consistently parse each entity, along with its complex attributes, affiliations, and inter-entity interactions, which are vital for accurate grounding. These examples demonstrate ChatGPT's superior robustness compared to ReCLIP's language parsing method, especially in challenging scenarios.

\clearpage
{
    \small
    \bibliographystyle{ieeenat_fullname}
    \bibliography{main,main2}

\begin{thebibliography}{79}
\providecommand{\natexlab}[1]{#1}
\providecommand{\url}[1]{\texttt{#1}}
\expandafter\ifx\csname urlstyle\endcsname\relax
  \providecommand{\doi}[1]{doi: #1}\else
  \providecommand{\doi}{doi: \begingroup \urlstyle{rm}\Url}\fi

\bibitem[Belanger and McCallum(2016)]{belanger2016structured}
David Belanger and Andrew McCallum.
\newblock Structured prediction energy networks.
\newblock In \emph{International Conference on Machine Learning}, pages 983--992. PMLR, 2016.

\bibitem[Carreras and M{\`a}rquez(2005)]{srl2}
Xavier Carreras and Llu{\'\i}s M{\`a}rquez.
\newblock Introduction to the conll-2005 shared task: Semantic role labeling.
\newblock In \emph{Proceedings of the ninth conference on computational natural language learning (CoNLL-2005)}, pages 152--164, 2005.

\bibitem[Chao et~al.(2015)Chao, Wang, He, Wang, and Deng]{hico}
Yu-Wei Chao, Zhan Wang, Yugeng He, Jiaxuan Wang, and Jia Deng.
\newblock Hico: A benchmark for recognizing human-object interactions in images.
\newblock In \emph{Proceedings of the IEEE international conference on computer vision}, pages 1017--1025, 2015.

\bibitem[Chen et~al.(2020)Chen, Li, Yu, El~Kholy, Ahmed, Gan, Cheng, and Liu]{Uniter}
Yen-Chun Chen, Linjie Li, Licheng Yu, Ahmed El~Kholy, Faisal Ahmed, Zhe Gan, Yu Cheng, and Jingjing Liu.
\newblock Uniter: Universal image-text representation learning.
\newblock In \emph{European conference on computer vision}, pages 104--120. Springer, 2020.

\bibitem[Christou and Tsoumakas(2021)]{christou2021improving}
Despina Christou and Grigorios Tsoumakas.
\newblock Improving distantly-supervised relation extraction through bert-based label and instance embeddings.
\newblock \emph{IEEE Access}, 9:\penalty0 62574--62582, 2021.

\bibitem[Cui et~al.(2021)Cui, Khandelwal, Artzi, Snavely, and Averbuch-Elor]{whos-waldo}
Yuqing Cui, Apoorv Khandelwal, Yoav Artzi, Noah Snavely, and Hadar Averbuch-Elor.
\newblock Who's waldo? linking people across text and images.
\newblock In \emph{Proceedings of the IEEE/CVF International Conference on Computer Vision}, pages 1374--1384, 2021.

\bibitem[Doveh et~al.(2023{\natexlab{a}})Doveh, Arbelle, Harary, Alfassy, Herzig, Kim, Giryes, Feris, Panda, Ullman, et~al.]{dac}
Sivan Doveh, Assaf Arbelle, Sivan Harary, Amit Alfassy, Roei Herzig, Donghyun Kim, Raja Giryes, Rogerio Feris, Rameswar Panda, Shimon Ullman, et~al.
\newblock Dense and aligned captions (dac) promote compositional reasoning in vl models.
\newblock \emph{arXiv preprint arXiv:2305.19595}, 2023{\natexlab{a}}.

\bibitem[Doveh et~al.(2023{\natexlab{b}})Doveh, Arbelle, Harary, Schwartz, Herzig, Giryes, Feris, Panda, Ullman, and Karlinsky]{structured-vl}
Sivan Doveh, Assaf Arbelle, Sivan Harary, Eli Schwartz, Roei Herzig, Raja Giryes, Rogerio Feris, Rameswar Panda, Shimon Ullman, and Leonid Karlinsky.
\newblock Teaching structured vision \& language concepts to vision \& language models.
\newblock In \emph{Proceedings of the IEEE/CVF Conference on Computer Vision and Pattern Recognition}, pages 2657--2668, 2023{\natexlab{b}}.

\bibitem[Fan et~al.(2024)Fan, Gu, Zhou, Yan, Jiang, Kuo, Guan, and Wang]{fan2024muffin}
Yue Fan, Jing Gu, Kaiwen Zhou, Qianqi Yan, Shan Jiang, Ching-Chen Kuo, Xinze Guan, and Xin~Eric Wang.
\newblock Muffin or chihuahua? challenging large vision-language models with multipanel vqa.
\newblock \emph{arXiv preprint arXiv:2401.15847}, 2024.

\bibitem[Finkel and Manning(2009)]{entity-recognition1}
Jenny~Rose Finkel and Christopher~D Manning.
\newblock Nested named entity recognition.
\newblock In \emph{Proceedings of the 2009 conference on empirical methods in natural language processing}, pages 141--150, 2009.

\bibitem[Gao et~al.(2020)Gao, Xu, Zou, and Huang]{hoi_drg20}
Chen Gao, Jiarui Xu, Yuliang Zou, and Jia-Bin Huang.
\newblock {DRG}: Dual relation graph for human-object interaction detection.
\newblock In \emph{ECCV}, 2020.

\bibitem[Gu and Yu(2020)]{gu2020data}
Jing Gu and Zhou Yu.
\newblock Data annealing for informal language understanding tasks.
\newblock In \emph{Findings of the Association for Computational Linguistics: EMNLP 2020}, pages 3153--3159, 2020.

\bibitem[Gu et~al.(2022)Gu, Stefani, Wu, Thomason, and Wang]{vision-navigation}
Jing Gu, Eliana Stefani, Qi Wu, Jesse Thomason, and Xin~Eric Wang.
\newblock Vision-and-language navigation: A survey of tasks, methods, and future directions.
\newblock \emph{arXiv preprint arXiv:2203.12667}, 2022.

\bibitem[Gupta and Kembhavi(2023)]{visual-programming}
Tanmay Gupta and Aniruddha Kembhavi.
\newblock Visual programming: Compositional visual reasoning without training.
\newblock In \emph{Proceedings of the IEEE/CVF Conference on Computer Vision and Pattern Recognition}, pages 14953--14962, 2023.

\bibitem[Gupta et~al.(2020)Gupta, Vahdat, Chechik, Yang, Kautz, and Hoiem]{gupta2020contrastive}
Tanmay Gupta, Arash Vahdat, Gal Chechik, Xiaodong Yang, Jan Kautz, and Derek Hoiem.
\newblock Contrastive learning for weakly supervised phrase grounding.
\newblock In \emph{European Conference on Computer Vision}, pages 752--768. Springer, 2020.

\bibitem[Han et~al.(2018)Han, Zhu, Yu, Wang, Yao, Liu, and Sun]{relationcls2}
Xu Han, Hao Zhu, Pengfei Yu, Ziyun Wang, Yuan Yao, Zhiyuan Liu, and Maosong Sun.
\newblock Fewrel: A large-scale supervised few-shot relation classification dataset with state-of-the-art evaluation.
\newblock \emph{arXiv preprint arXiv:1810.10147}, 2018.

\bibitem[Han et~al.(2024)Han, Gao, Liu, Zhang, et~al.]{peft_survey}
Zeyu Han, Chao Gao, Jinyang Liu, Sai~Qian Zhang, et~al.
\newblock Parameter-efficient fine-tuning for large models: A comprehensive survey.
\newblock \emph{arXiv preprint arXiv:2403.14608}, 2024.

\bibitem[He et~al.(2017)He, Lee, Lewis, and Zettlemoyer]{srl1}
Luheng He, Kenton Lee, Mike Lewis, and Luke Zettlemoyer.
\newblock Deep semantic role labeling: What works and what’s next.
\newblock In \emph{Proceedings of the 55th Annual Meeting of the Association for Computational Linguistics (Volume 1: Long Papers)}, pages 473--483, 2017.

\bibitem[Herzig et~al.(2023)Herzig, Mendelson, Karlinsky, Arbelle, Feris, Darrell, and Globerson]{clipscenegraph}
Roei Herzig, Alon Mendelson, Leonid Karlinsky, Assaf Arbelle, Rogerio Feris, Trevor Darrell, and Amir Globerson.
\newblock Incorporating structured representations into pretrained vision \& language models using scene graphs.
\newblock \emph{arXiv preprint arXiv:2305.06343}, 2023.

\bibitem[Hossain et~al.(2019)Hossain, Sohel, Shiratuddin, and Laga]{captionsurvey1}
MD~Zakir Hossain, Ferdous Sohel, Mohd~Fairuz Shiratuddin, and Hamid Laga.
\newblock A comprehensive survey of deep learning for image captioning.
\newblock \emph{ACM Computing Surveys (CsUR)}, 51\penalty0 (6):\penalty0 1--36, 2019.

\bibitem[Hu et~al.(2021)Hu, Shen, Wallis, Allen-Zhu, Li, Wang, Wang, and Chen]{lora}
Edward~J Hu, Yelong Shen, Phillip Wallis, Zeyuan Allen-Zhu, Yuanzhi Li, Shean Wang, Lu Wang, and Weizhu Chen.
\newblock Lora: Low-rank adaptation of large language models.
\newblock \emph{arXiv preprint arXiv:2106.09685}, 2021.

\bibitem[Huang et~al.(2023)Huang, Tang, Chen, Zhang, Zhang, Chen, Zhao, Lv, Hu, and Zhang]{structure-CLIP}
Yufeng Huang, Jiji Tang, Zhuo Chen, Rongsheng Zhang, Xinfeng Zhang, Weijie Chen, Zeng Zhao, Tangjie Lv, Zhipeng Hu, and Wen Zhang.
\newblock Structure-clip: Enhance multi-modal language representations with structure knowledge.
\newblock \emph{arXiv preprint arXiv:2305.06152}, 2023.

\bibitem[Ji and Grishman(2008)]{eventext1}
Heng Ji and Ralph Grishman.
\newblock Refining event extraction through cross-document inference.
\newblock In \emph{Proceedings of ACL-08: Hlt}, pages 254--262, 2008.

\bibitem[Jiang et~al.(2022)Jiang, He, Xu, and Wang]{comclip}
Kenan Jiang, Xuehai He, Ruize Xu, and Xin~Eric Wang.
\newblock Comclip: Training-free compositional image and text matching.
\newblock \emph{arXiv preprint arXiv:2211.13854}, 2022.

\bibitem[Kamath et~al.(2021)Kamath, Singh, LeCun, Synnaeve, Misra, and Carion]{mdetr}
Aishwarya Kamath, Mannat Singh, Yann LeCun, Gabriel Synnaeve, Ishan Misra, and Nicolas Carion.
\newblock Mdetr-modulated detection for end-to-end multi-modal understanding.
\newblock In \emph{Proceedings of the IEEE/CVF International Conference on Computer Vision}, pages 1780--1790, 2021.

\bibitem[Kim et~al.(2021)Kim, Lee, Kang, Kim, and Kim]{hoi_hotr21}
Bumsoo Kim, Junhyun Lee, Jaewoo Kang, Eun-Sol Kim, and Hyunwoo~J Kim.
\newblock {HOTR}: End-to-end human-object interaction detection with transformers.
\newblock In \emph{CVPR}, 2021.

\bibitem[Kim et~al.(2023)Kim, Jung, and Cho]{hoi_muren23}
Sanghyun Kim, Deunsol Jung, and Minsu Cho.
\newblock Relational context learning for human-object interaction detection.
\newblock In \emph{CVPR}, 2023.

\bibitem[Kirillov et~al.(2023)Kirillov, Mintun, Ravi, Mao, Rolland, Gustafson, Xiao, Whitehead, Berg, Lo, et~al.]{kirillov2023segment}
Alexander Kirillov, Eric Mintun, Nikhila Ravi, Hanzi Mao, Chloe Rolland, Laura Gustafson, Tete Xiao, Spencer Whitehead, Alexander~C Berg, Wan-Yen Lo, et~al.
\newblock Segment anything.
\newblock In \emph{Proceedings of the IEEE/CVF International Conference on Computer Vision}, pages 4015--4026, 2023.

\bibitem[Krishna et~al.(2017)Krishna, Zhu, Groth, Johnson, Hata, Kravitz, Chen, Kalantidis, Li, Shamma, et~al.]{vg}
Ranjay Krishna, Yuke Zhu, Oliver Groth, Justin Johnson, Kenji Hata, Joshua Kravitz, Stephanie Chen, Yannis Kalantidis, Li-Jia Li, David~A Shamma, et~al.
\newblock Visual genome: Connecting language and vision using crowdsourced dense image annotations.
\newblock \emph{International journal of computer vision}, 123:\penalty0 32--73, 2017.

\bibitem[Krishna et~al.(2018)Krishna, Chami, Bernstein, and Fei-Fei]{visual-relationships}
Ranjay Krishna, Ines Chami, Michael Bernstein, and Li Fei-Fei.
\newblock Referring relationships.
\newblock In \emph{Proceedings of the IEEE conference on computer vision and pattern recognition}, pages 6867--6876, 2018.

\bibitem[Lee et~al.(2018)Lee, He, and Zettlemoyer]{corefres1}
Kenton Lee, Luheng He, and Luke Zettlemoyer.
\newblock Higher-order coreference resolution with coarse-to-fine inference.
\newblock \emph{arXiv preprint arXiv:1804.05392}, 2018.

\bibitem[Li et~al.(2022)Li, Shakhnarovich, and Yeh]{clip-phrase-grounding}
Jiahao Li, Greg Shakhnarovich, and Raymond~A Yeh.
\newblock Adapting clip for phrase localization without further training.
\newblock \emph{arXiv preprint arXiv:2204.03647}, 2022.

\bibitem[Li et~al.(2013)Li, Ji, and Huang]{eventext2}
Qi Li, Heng Ji, and Liang Huang.
\newblock Joint event extraction via structured prediction with global features.
\newblock In \emph{Proceedings of the 51st Annual Meeting of the Association for Computational Linguistics (Volume 1: Long Papers)}, pages 73--82, 2013.

\bibitem[Li et~al.(2019)Li, Feng, Meng, Han, Wu, and Li]{entity-recognition2}
Xiaoya Li, Jingrong Feng, Yuxian Meng, Qinghong Han, Fei Wu, and Jiwei Li.
\newblock A unified mrc framework for named entity recognition.
\newblock \emph{arXiv preprint arXiv:1910.11476}, 2019.

\bibitem[Li et~al.(2017)Li, Ouyang, Zhou, Wang, and Wang]{li2017scene}
Yikang Li, Wanli Ouyang, Bolei Zhou, Kun Wang, and Xiaogang Wang.
\newblock Scene graph generation from objects, phrases and region captions.
\newblock In \emph{ICCV}, 2017.

\bibitem[Liao et~al.(2022)Liao, Zhang, Lu, Wang, Li, and Liu]{hoi_genvlkt22}
Yue Liao, Aixi Zhang, Miao Lu, Yongliang Wang, Xiaobo Li, and Si Liu.
\newblock Gen-vlkt: Simplify association and enhance interaction understanding for hoi detection.
\newblock In \emph{CVPR}, 2022.

\bibitem[Lin et~al.(2023)Lin, Yuan, Wu, Wang, and Wang]{uninext}
Fangjian Lin, Jianlong Yuan, Sitong Wu, Fan Wang, and Zhibin Wang.
\newblock Uninext: Exploring a unified architecture for vision recognition.
\newblock \emph{arXiv preprint arXiv:2304.13700}, 2023.

\bibitem[Lin et~al.(2014{\natexlab{a}})Lin, Maire, Belongie, Hays, Perona, Ramanan, Doll{\'a}r, and Zitnick]{coco}
Tsung-Yi Lin, Michael Maire, Serge Belongie, James Hays, Pietro Perona, Deva Ramanan, Piotr Doll{\'a}r, and C~Lawrence Zitnick.
\newblock Microsoft coco: Common objects in context.
\newblock In \emph{ECCV}, 2014{\natexlab{a}}.

\bibitem[Lin et~al.(2014{\natexlab{b}})Lin, Maire, Belongie, Hays, Perona, Ramanan, Doll{\'a}r, and Zitnick]{mscoco}
Tsung-Yi Lin, Michael Maire, Serge Belongie, James Hays, Pietro Perona, Deva Ramanan, Piotr Doll{\'a}r, and C~Lawrence Zitnick.
\newblock Microsoft coco: Common objects in context.
\newblock In \emph{Computer Vision--ECCV 2014: 13th European Conference, Zurich, Switzerland, September 6-12, 2014, Proceedings, Part V 13}, pages 740--755. Springer, 2014{\natexlab{b}}.

\bibitem[Liu et~al.(2023)Liu, Huang, Kang, Chen, and Wang]{diffvg}
Xuyang Liu, Siteng Huang, Yachen Kang, Honggang Chen, and Donglin Wang.
\newblock Vgdiffzero: Text-to-image diffusion models can be zero-shot visual grounders.
\newblock \emph{arXiv preprint arXiv:2309.01141}, 2023.

\bibitem[Liu et~al.(2020)Liu, Yuan, and Chen]{liu2020consnet}
Ye Liu, Junsong Yuan, and Chang~Wen Chen.
\newblock Consnet: Learning consistency graph for zero-shot human-object interaction detection.
\newblock In \emph{Proceedings of the 28th ACM International Conference on Multimedia}, 2020.

\bibitem[Ma et~al.(2023)Ma, Wang, Wang, and Wei]{hoi_fgahoi23}
Shuailei Ma, Yuefeng Wang, Shanze Wang, and Ying Wei.
\newblock Fgahoi: Fine-grained anchors for human-object interaction detection.
\newblock \emph{arXiv preprint arXiv:2301.04019}, 2023.

\bibitem[Mangrulkar et~al.(2022)Mangrulkar, Gugger, Debut, Belkada, Paul, and Bossan]{peft}
Sourab Mangrulkar, Sylvain Gugger, Lysandre Debut, Younes Belkada, Sayak Paul, and Benjamin Bossan.
\newblock Peft: State-of-the-art parameter-efficient fine-tuning methods.
\newblock \url{https://github.com/huggingface/peft}, 2022.

\bibitem[Mao et~al.(2016)Mao, Huang, Toshev, Camburu, Yuille, and Murphy]{refcoco1}
Junhua Mao, Jonathan Huang, Alexander Toshev, Oana Camburu, Alan~L Yuille, and Kevin Murphy.
\newblock Generation and comprehension of unambiguous object descriptions.
\newblock In \emph{Proceedings of the IEEE conference on computer vision and pattern recognition}, pages 11--20, 2016.

\bibitem[Marino et~al.(2019)Marino, Rastegari, Farhadi, and Mottaghi]{vqa1}
Kenneth Marino, Mohammad Rastegari, Ali Farhadi, and Roozbeh Mottaghi.
\newblock Ok-vqa: A visual question answering benchmark requiring external knowledge.
\newblock In \emph{Proceedings of the IEEE/cvf conference on computer vision and pattern recognition}, pages 3195--3204, 2019.

\bibitem[Paolini et~al.(2021)Paolini, Athiwaratkun, Krone, Ma, Achille, Anubhai, Santos, Xiang, and Soatto]{paolini2021structured}
Giovanni Paolini, Ben Athiwaratkun, Jason Krone, Jie Ma, Alessandro Achille, Rishita Anubhai, Cicero Nogueira~dos Santos, Bing Xiang, and Stefano Soatto.
\newblock Structured prediction as translation between augmented natural languages.
\newblock In \emph{ICLR}, 2021.

\bibitem[Plummer et~al.(2015)Plummer, Wang, Cervantes, Caicedo, Hockenmaier, and Lazebnik]{flickr30k}
Bryan~A Plummer, Liwei Wang, Chris~M Cervantes, Juan~C Caicedo, Julia Hockenmaier, and Svetlana Lazebnik.
\newblock Flickr30k entities: Collecting region-to-phrase correspondences for richer image-to-sentence models.
\newblock In \emph{Proceedings of the IEEE international conference on computer vision}, pages 2641--2649, 2015.

\bibitem[Pratt et~al.(2020)Pratt, Yatskar, Weihs, Farhadi, and Kembhavi]{swig}
Sarah Pratt, Mark Yatskar, Luca Weihs, Ali Farhadi, and Aniruddha Kembhavi.
\newblock Grounded situation recognition.
\newblock In \emph{Computer Vision--ECCV 2020: 16th European Conference, Glasgow, UK, August 23--28, 2020, Proceedings, Part IV 16}, pages 314--332. Springer, 2020.

\bibitem[Radford et~al.(2021)Radford, Kim, Hallacy, Ramesh, Goh, Agarwal, Sastry, Askell, Mishkin, Clark, et~al.]{clip}
Alec Radford, Jong~Wook Kim, Chris Hallacy, Aditya Ramesh, Gabriel Goh, Sandhini Agarwal, Girish Sastry, Amanda Askell, Pamela Mishkin, Jack Clark, et~al.
\newblock Learning transferable visual models from natural language supervision.
\newblock In \emph{International conference on machine learning}, pages 8748--8763. PMLR, 2021.

\bibitem[Selvaraju et~al.(2017)Selvaraju, Cogswell, Das, Vedantam, Parikh, and Batra]{gradcam}
Ramprasaath~R Selvaraju, Michael Cogswell, Abhishek Das, Ramakrishna Vedantam, Devi Parikh, and Dhruv Batra.
\newblock Grad-cam: Visual explanations from deep networks via gradient-based localization.
\newblock In \emph{Proceedings of the IEEE international conference on computer vision}, pages 618--626, 2017.

\bibitem[Shao et~al.(2019)Shao, Li, Zhang, Peng, Yu, Zhang, Li, and Sun]{object365}
Shuai Shao, Zeming Li, Tianyuan Zhang, Chao Peng, Gang Yu, Xiangyu Zhang, Jing Li, and Jian Sun.
\newblock Objects365: A large-scale, high-quality dataset for object detection.
\newblock In \emph{Proceedings of the IEEE/CVF international conference on computer vision}, pages 8430--8439, 2019.

\bibitem[Shi and Lin(2019)]{bertsrl}
Peng Shi and Jimmy Lin.
\newblock Simple bert models for relation extraction and semantic role labeling.
\newblock \emph{arXiv preprint arXiv:1904.05255}, 2019.

\bibitem[Shibuya and Hovy(2020)]{entity-recognition3}
Takashi Shibuya and Eduard Hovy.
\newblock Nested named entity recognition via second-best sequence learning and decoding.
\newblock \emph{Transactions of the Association for Computational Linguistics}, 8:\penalty0 605--620, 2020.

\bibitem[Singh et~al.(2022)Singh, Hu, Goswami, Couairon, Galuba, Rohrbach, and Kiela]{flava}
Amanpreet Singh, Ronghang Hu, Vedanuj Goswami, Guillaume Couairon, Wojciech Galuba, Marcus Rohrbach, and Douwe Kiela.
\newblock Flava: A foundational language and vision alignment model.
\newblock In \emph{Proceedings of the IEEE/CVF Conference on Computer Vision and Pattern Recognition}, pages 15638--15650, 2022.

\bibitem[Stefanini et~al.(2022)Stefanini, Cornia, Baraldi, Cascianelli, Fiameni, and Cucchiara]{captionsurvey2}
Matteo Stefanini, Marcella Cornia, Lorenzo Baraldi, Silvia Cascianelli, Giuseppe Fiameni, and Rita Cucchiara.
\newblock From show to tell: A survey on deep learning-based image captioning.
\newblock \emph{IEEE transactions on pattern analysis and machine intelligence}, 45\penalty0 (1):\penalty0 539--559, 2022.

\bibitem[Subramanian et~al.(2022)Subramanian, Merrill, Darrell, Gardner, Singh, and Rohrbach]{reclip}
Sanjay Subramanian, William Merrill, Trevor Darrell, Matt Gardner, Sameer Singh, and Anna Rohrbach.
\newblock Reclip: A strong zero-shot baseline for referring expression comprehension.
\newblock In \emph{ACL}, 2022.

\bibitem[Tang et~al.(2019)Tang, Zhang, Wu, Luo, and Liu]{tang2019learning}
Kaihua Tang, Hanwang Zhang, Baoyuan Wu, Wenhan Luo, and Wei Liu.
\newblock Learning to compose dynamic tree structures for visual contexts.
\newblock In \emph{CVPR}, 2019.

\bibitem[Wei et~al.(2023)Wei, Wei, Tay, Tran, Webson, Lu, Chen, Liu, Huang, Zhou, et~al.]{incontext}
Jerry Wei, Jason Wei, Yi Tay, Dustin Tran, Albert Webson, Yifeng Lu, Xinyun Chen, Hanxiao Liu, Da Huang, Denny Zhou, et~al.
\newblock Larger language models do in-context learning differently.
\newblock \emph{arXiv preprint arXiv:2303.03846}, 2023.

\bibitem[Wu et~al.(2023)Wu, Zhang, Jin, Liu, and Loy]{bag-of-region}
Size Wu, Wenwei Zhang, Sheng Jin, Wentao Liu, and Chen~Change Loy.
\newblock Aligning bag of regions for open-vocabulary object detection.
\newblock In \emph{Proceedings of the IEEE/CVF Conference on Computer Vision and Pattern Recognition}, pages 15254--15264, 2023.

\bibitem[Wu et~al.(2020)Wu, Wang, Yuan, Wu, and Li]{corefres2}
Wei Wu, Fei Wang, Arianna Yuan, Fei Wu, and Jiwei Li.
\newblock Corefqa: Coreference resolution as query-based span prediction.
\newblock In \emph{Proceedings of the 58th Annual Meeting of the Association for Computational Linguistics}, pages 6953--6963, 2020.

\bibitem[Xu et~al.(2023)Xu, Liu, Shen, Han, Li, Yue, Peng, Liu, Yao, and Xu]{gentopia}
Binfeng Xu, Xukun Liu, Hua Shen, Zeyu Han, Yuhan Li, Murong Yue, Zhiyuan Peng, Yuchen Liu, Ziyu Yao, and Dongkuan Xu.
\newblock Gentopia: A collaborative platform for tool-augmented llms.
\newblock \emph{arXiv preprint arXiv:2308.04030}, 2023.

\bibitem[Xu et~al.(2017)Xu, Zhu, Choy, and Fei-Fei]{xu2017scene}
Danfei Xu, Yuke Zhu, Christopher~B Choy, and Li Fei-Fei.
\newblock Scene graph generation by iterative message passing.
\newblock In \emph{CVPR}, 2017.

\bibitem[Yang et~al.(2022)Yang, Ang, Guo, Zhou, Zhang, and Liu]{psg}
Jingkang Yang, Yi~Zhe Ang, Zujin Guo, Kaiyang Zhou, Wayne Zhang, and Ziwei Liu.
\newblock Panoptic scene graph generation.
\newblock In \emph{European Conference on Computer Vision}, pages 178--196. Springer, 2022.

\bibitem[Yao et~al.(2021)Yao, Zhang, Zhang, Liu, Chua, and Sun]{cpt}
Yuan Yao, Ao Zhang, Zhengyan Zhang, Zhiyuan Liu, Tat-Seng Chua, and Maosong Sun.
\newblock Cpt: Colorful prompt tuning for pre-trained vision-language models.
\newblock \emph{arXiv preprint arXiv:2109.11797}, 2021.

\bibitem[Yu et~al.(2016)Yu, Poirson, Yang, Berg, and Berg]{refcoco2}
Licheng Yu, Patrick Poirson, Shan Yang, Alexander~C Berg, and Tamara~L Berg.
\newblock Modeling context in referring expressions.
\newblock In \emph{Computer Vision--ECCV 2016: 14th European Conference, Amsterdam, The Netherlands, October 11-14, 2016, Proceedings, Part II 14}, pages 69--85. Springer, 2016.

\bibitem[Yu et~al.(2018)Yu, Lin, Shen, Yang, Lu, Bansal, and Berg]{mattnet}
Licheng Yu, Zhe Lin, Xiaohui Shen, Jimei Yang, Xin Lu, Mohit Bansal, and Tamara~L Berg.
\newblock Mattnet: Modular attention network for referring expression comprehension.
\newblock In \emph{Proceedings of the IEEE conference on computer vision and pattern recognition}, pages 1307--1315, 2018.

\bibitem[Yu et~al.(2023)Yu, Seo, and Son]{zero-shot-seg}
Seonghoon Yu, Paul~Hongsuck Seo, and Jeany Son.
\newblock Zero-shot referring image segmentation with global-local context features.
\newblock In \emph{Proceedings of the IEEE/CVF Conference on Computer Vision and Pattern Recognition}, pages 19456--19465, 2023.

\bibitem[Yu et~al.(2019)Yu, Yu, Cui, Tao, and Tian]{vqa2}
Zhou Yu, Jun Yu, Yuhao Cui, Dacheng Tao, and Qi Tian.
\newblock Deep modular co-attention networks for visual question answering.
\newblock In \emph{Proceedings of the IEEE/CVF conference on computer vision and pattern recognition}, pages 6281--6290, 2019.

\bibitem[Yuksekgonul et~al.(2022)Yuksekgonul, Bianchi, Kalluri, Jurafsky, and Zou]{bows}
Mert Yuksekgonul, Federico Bianchi, Pratyusha Kalluri, Dan Jurafsky, and James Zou.
\newblock When and why vision-language models behave like bags-of-words, and what to do about it?
\newblock In \emph{The Eleventh International Conference on Learning Representations}, 2022.

\bibitem[Zellers et~al.(2018)Zellers, Yatskar, Thomson, and Choi]{zellers2018neural}
Rowan Zellers, Mark Yatskar, Sam Thomson, and Yejin Choi.
\newblock Neural motifs: Scene graph parsing with global context.
\newblock In \emph{CVPR}, 2018.

\bibitem[Zhang et~al.(2021{\natexlab{a}})Zhang, Liao, Liu, Lu, Wang, Gao, and Li]{hoi_cdn21}
Aixi Zhang, Yue Liao, Si Liu, Miao Lu, Yongliang Wang, Chen Gao, and Xiaobo Li.
\newblock Mining the benefits of two-stage and one-stage hoi detection.
\newblock In \emph{NeurIPS}, 2021{\natexlab{a}}.

\bibitem[Zhang et~al.(2021{\natexlab{b}})Zhang, Campbell, and Gould]{hoi_scg21}
Frederic~Z Zhang, Dylan Campbell, and Stephen Gould.
\newblock Spatially conditioned graphs for detecting human-object interactions.
\newblock In \emph{ICCV}, 2021{\natexlab{b}}.

\bibitem[Zhang et~al.(2022{\natexlab{a}})Zhang, Campbell, and Gould]{hoi_upt22}
Frederic~Z Zhang, Dylan Campbell, and Stephen Gould.
\newblock Efficient two-stage detection of human-object interactions with a novel unary-pairwise transformer.
\newblock In \emph{CVPR}, 2022{\natexlab{a}}.

\bibitem[Zhang et~al.(2017{\natexlab{a}})Zhang, Kyaw, Yu, and Chang]{zhang2017ppr}
Hanwang Zhang, Zawlin Kyaw, Jinyang Yu, and Shih-Fu Chang.
\newblock Ppr-fcn: Weakly supervised visual relation detection via parallel pairwise r-fcn.
\newblock In \emph{ICCV}, 2017{\natexlab{a}}.

\bibitem[Zhang et~al.(2022{\natexlab{b}})Zhang, Zhang, Hu, Chen, Li, Dai, Wang, Yuan, Hwang, and Gao]{glipv2}
Haotian Zhang, Pengchuan Zhang, Xiaowei Hu, Yen-Chun Chen, Liunian Li, Xiyang Dai, Lijuan Wang, Lu Yuan, Jenq-Neng Hwang, and Jianfeng Gao.
\newblock Glipv2: Unifying localization and vision-language understanding.
\newblock \emph{Advances in Neural Information Processing Systems}, 35:\penalty0 36067--36080, 2022{\natexlab{b}}.

\bibitem[Zhang et~al.(2017{\natexlab{b}})Zhang, Zhong, Chen, Angeli, and Manning]{relationcls1}
Yuhao Zhang, Victor Zhong, Danqi Chen, Gabor Angeli, and Christopher~D Manning.
\newblock Position-aware attention and supervised data improve slot filling.
\newblock In \emph{Conference on Empirical Methods in Natural Language Processing}, 2017{\natexlab{b}}.

\bibitem[Zhang et~al.(2022{\natexlab{c}})Zhang, Pan, Yao, Huang, Mei, and Chen]{hoi_stip22}
Yong Zhang, Yingwei Pan, Ting Yao, Rui Huang, Tao Mei, and Chang-Wen Chen.
\newblock Exploring structure-aware transformer over interaction proposals for human-object interaction detection.
\newblock In \emph{CVPR}, 2022{\natexlab{c}}.

\bibitem[Zhong et~al.(2021)Zhong, Shi, Yang, Xu, and Li]{zhong2021learning}
Yiwu Zhong, Jing Shi, Jianwei Yang, Chenliang Xu, and Yin Li.
\newblock Learning to generate scene graph from natural language supervision.
\newblock In \emph{ICCV}, 2021.

\bibitem[Zhu et~al.(2023)Zhu, Xie, Xie, and Jiang]{zhu2023diagnosing}
Fangrui Zhu, Yiming Xie, Weidi Xie, and Huaizu Jiang.
\newblock Diagnosing human-object interaction detectors.
\newblock \emph{arXiv preprint arXiv:2308.08529}, 2023.

\end{thebibliography}
}
\end{document}